\documentclass[10pt,twocolumn,letterpaper]{article}
\usepackage{paperstyle}

\usepackage{booktabs}
\usepackage{amsmath,amssymb}
\usepackage{pifont}
\usepackage{placeins}
\usepackage{cuted}
\usepackage{caption}

\newcommand{\cmark}{\textcolor{ForestGreen}{\ding{51}}}
\newcommand{\xmark}{\textcolor{BrickRed}{\ding{55}}}
\graphicspath{{figures/}}
\usepackage{xspace}
\newcommand{\proposed}{SWEEP-Bench\xspace}

\definecolor{linkblue}{rgb}{0.21,0.49,0.74}
\usepackage[pagebackref,breaklinks,colorlinks,allcolors=linkblue]{hyperref}

\title{\proposed: Do Embodied Policies Act on the Speed They See?}

\renewcommand{\topfraction}{0.9}
\renewcommand{\textfraction}{0.07}
\renewcommand{\floatpagefraction}{0.85}

\author{Hiroaki Honda \quad Koya Sakamoto \quad Shu Morikuni \quad Yusuke Iwasawa \quad Taiki Miyanishi\\
The University of Tokyo, Japan
}

\begin{document}
\maketitle
\begin{strip}
  \centering
  \includegraphics[width=\textwidth]{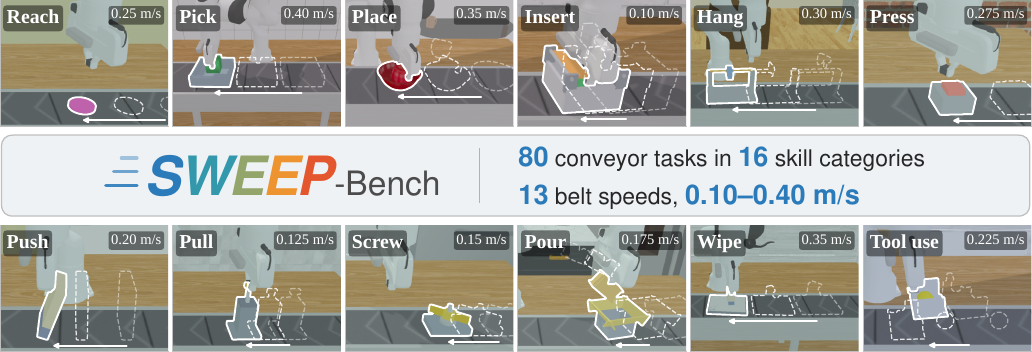}
  \captionof{figure}{%
    \textbf{\proposed: 80 conveyor tasks in 16 skill categories,
    at 13 belt speeds.}
    One task from each of 12 categories, each at a belt speed drawn at random from the 13 (top right).
    Solid outlines show the task's objects now and dashed ones 0.5\,s and 1\,s earlier (fainter with age); arrows show their motion over the last 1\,s.
  }
  \label{fig:teaser}
\end{strip}

\begin{abstract}

Robot policies must adapt to the speed of moving objects, yet existing benchmarks rarely measure this ability across diverse tasks and speeds.
We introduce \proposed, a simulation benchmark with 80 conveyor-belt manipulation tasks spanning 16 skill categories, each evaluated at 13 belt speeds (0.10--0.40~m/s).
Objects are transported through physical contact with a visibly moving belt.
Each task has a static counterpart, and each evaluation episode uses the same randomized scene at all 13 speeds.
We fine-tune eight vision-language-action and world-action models on demonstrations spanning the full speed range, comparing observations from the current frame alone with those including two past frames.
For six models, the success rate peaks at intermediate speeds, and their failures indicate that they tend to produce actions suited to the average demonstration speed rather than to the actual one.
Past frames raise the success rate by 10--40 percentage points for three models, and linear probes show that the belt speed is also most linearly decodable from these three.
Adding ground-truth speed input to four of the models improves all of them, and a speed estimated from the images keeps most of this gain.
These results show that \proposed exposes speed-dependent failures and suggest that past frames help only policies from which the speed is most decodable, such as those pretrained on multiple frames.
Project page: \href{https://damin3927.github.io/SWEEP-Bench}{damin3927.github.io/SWEEP-Bench}
\end{abstract}

\section{Introduction}
\label{sec:intro}

Robot policies trained from demonstrations are largely evaluated on benchmarks in which objects remain stationary until robot contact~\cite{liu2023libero,nasiriany2024robocasa}.
Manipulating a moving object requires the robot to adjust the timing and location of contact to the object's speed.
Recent studies report that vision-language-action (VLA) models struggle with moving objects~\cite{xie2026dynamicvla} and examine temporal motion cues~\cite{fang2026domino}, inference latency~\cite{wen2026vlalagbench}, and physical prediction~\cite{yun2026physmani}.
However, it remains unclear whether these policies adapt their actions to object speed when that speed can be inferred from past frames.

Answering this question requires evaluating success rates across speeds to reveal speed-dependent failures and across diverse manipulation skills so that the findings are not tied to a few tasks.
Each task must also be solvable at the tested speeds so that failures reflect policy limitations rather than task infeasibility.
Existing moving-object benchmarks rarely provide all three.
Those with larger task sets report success rates at no more than three speeds or provide no per-speed results; those reporting more speeds do so for at most four tasks.
Few also check during task construction that an expert solves every task at several speeds.

We introduce \proposed (Speed-Wise Evaluation of Embodied Policies), a simulation benchmark for moving-object manipulation on a conveyor belt.
We use a conveyor because we can set its speed while keeping the rest of the scene fixed, it supports many manipulation skills, and its motion is visible in the camera images.
The benchmark contains 80 conveyor-belt tasks in 16 skill categories, each evaluated at 13 speeds from 0.10 to 0.40~m/s, with constant speeds in the main protocol and within-episode speed changes in an acceleration setting.
To isolate the effect of speed, each evaluation episode uses the same randomized scene at all 13 speeds, and the training demonstrations span the full speed range.
To ensure that failures reflect the policy, a scripted expert solves every task at four speeds spanning the range, and each task has a static counterpart that checks whether it can be learned with a stationary object.

We fine-tune six VLA models and two video-based world-action models~\cite{kim2026cosmospolicy,fastwam2026}.
For each model, we compare the current frame alone with the current frame plus two past frames and evaluate both policies on every task at every speed.
For six models, the success rate peaks at intermediate speeds with or without past frames, as the policies tend to produce actions suited to the average demonstration speed.
Measuring success at 13 speeds makes this pattern visible, whereas averaging over speeds hides it.

A small network on the optical flow between two frames 0.1~s apart estimates the belt speed with a mean absolute error of 0.010~m/s.
Past frames, however, raise the mean success rate by 10--40 percentage points for three models and by at most one point for the other five.
Linear probes on the fine-tuned policies show that the belt speed is most linearly decodable from these three ($R^2 = 0.86$--$0.99$ at the action module), whereas the other five reach at most $R^2 = 0.58$ and gain little from past frames.
The speed is useful once a policy has it: on a 20-task subset, adding ground-truth speed input to four of the models improves all of them, and an image-based speed adapter recovers most of this gain.
In 720 real-conveyor trials, past frames raise Fast-WAM's score slightly more than $\pi_{0.5}$'s, as in simulation, and adding belt speed input raises $\pi_{0.5}$'s score from 31 to 80, and to 66 when the adapter estimates the speed.

Our contributions are:
\begin{itemize}
\item We introduce \proposed, with 80 tasks across 16 skill categories and 13 speeds, paired evaluation scenes, an expert check of every task at four speeds, and a static counterpart for every task.
We will release the code, demonstrations, and evaluation protocol.
\item We evaluate eight VLA and world-action models with and without past frames on every task at every speed, showing that six of them peak at intermediate speeds and tend to produce actions suited to the average demonstration speed.
\item Using linear probes and speed-conditioned policies, we find that past frames help only the three models from which the belt speed is most linearly decodable, and that a speed estimated from the images keeps most of the gain from ground-truth speed input.
\end{itemize}

\begin{figure*}[t]
  \centering
  \includegraphics[width=\textwidth]{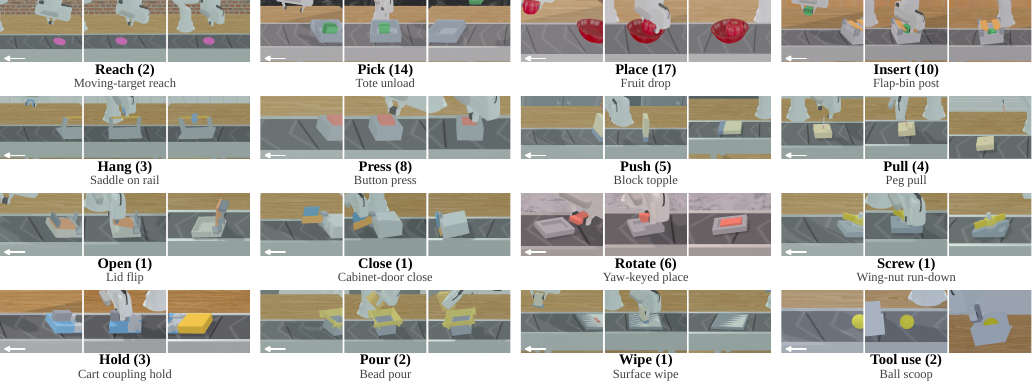}
\caption{\textbf{One task per skill category}
(16 categories; task counts in parentheses).
Triplets show scripted-expert episodes at 0.40\,m/s;
time runs left to right.
Crops follow the object; arrows show belt direction.}
\label{fig:tasks}
\end{figure*}

\section{Related Work}
\label{sec:relatedwork}

\begin{table}[t]
\centering
\caption{\textbf{Speed-wise coverage of dynamic manipulation benchmarks.}
\emph{Motion}: what moves the object; \emph{direct}: the object moves with no visible carrier.
\emph{Acceleration}: speeding-up or slowing-down motion is a separate evaluation condition.
\emph{Expert check}: task construction checks that an expert solves every task at several tested speeds.
\emph{Skill categories}: tasks grouped by the main action of their instruction; actions outside \proposed's 16 categories count separately.
\emph{Tasks} exclude speed, acceleration, direction, and difficulty variants.}
\label{tab:benchmarks}
\scriptsize
\setlength{\tabcolsep}{1pt}
\begin{tabular*}{\columnwidth}{@{\extracolsep{\fill}}lcccccrc@{}}
\toprule
Benchmark & Motion & \shortstack{Static\\counter-\\part} & \shortstack{Accel-\\eration} & \shortstack{Expert\\check} & \shortstack{Skill\\cate-\\gories} & Tasks & Speeds \\
\midrule
\multicolumn{8}{@{}l}{\emph{No per-speed success}} \\
LIBERO-Industry~\cite{liberoindustry2026} & belt & \xmark & \xmark & \xmark & 1 & 12 & 3 \\
DynaGrasp-32~\cite{chang2026dynawm} & direct & \xmark & \xmark & \xmark & 2 & 25 & range \\
ReflexBench~\cite{chen2026reflex} & mixed & \xmark & \xmark & \xmark & 5 & 6 & not reported \\
DOM~\cite{xie2026dynamicvla} & direct & \xmark & \xmark & \xmark & 1 & 1 & range \\
\midrule
\multicolumn{8}{@{}l}{\emph{Success reported per speed}} \\
VLALagBench~\cite{wen2026vlalagbench} & mixed & \xmark & \xmark & \xmark & 2 & 4 & 9 \\
DOMINO~\cite{fang2026domino} & direct & \cmark & \xmark & \xmark & 9 & 35 & 3 bands \\
PhysMani-Bench~\cite{yun2026physmani} & mixed & \xmark & \xmark & \xmark & 5 & 8 & 2 bands \\
MIKASA-Robo~\cite{cherepanov2025mikasa} & direct & \xmark & \xmark & \xmark & 2 & 2 & 3 bands \\
DGBench~\cite{burgesslimerick2022dgbench} & platform & \xmark & \xmark & \xmark & 1 & 1 & 5 \\
AHEAD~\cite{syed2026ahead} & mixed & \xmark & \cmark & \xmark & 4 & 15 & 4 \\
MoveBench~\cite{zhang2026ppc} & direct & \cmark & \cmark & \xmark & 1 & 1 & 3 bands \\
DQ-Bench~\cite{liang2026dqbench} & platform & \xmark & \xmark & \xmark & 1 & 1 & 2 bands \\
LIBERO-Dynamic~\cite{ren2026liberodynamic} & direct & \cmark & \xmark & \cmark & 5 & 40 & 3 \\
\midrule
\textbf{\proposed (ours)} & \textbf{belt} & \cmark & \cmark & \cmark & \textbf{16} & \textbf{80} & \textbf{13} \\
\bottomrule
\end{tabular*}
\end{table}

\paragraph{Benchmarks for moving-object manipulation.}
Dynamic grasping benchmarks such as DGBench~\cite{burgesslimerick2022dgbench} and DQ-Bench~\cite{liang2026dqbench} vary target speed for a single manipulation skill.
Larger suites broaden task coverage: DOMINO~\cite{fang2026domino} includes 35 dynamic tasks and static counterparts, DynaGrasp-32~\cite{chang2026dynawm} contains 25 moving-object tasks, LIBERO-Industry~\cite{liberoindustry2026} covers industrial transfer workflows, and LIBERO-Dynamic~\cite{ren2026liberodynamic} evaluates 40 LIBERO tasks at three speeds.
These studies do not report dense speed-wise success-rate curves across their full task sets.
Benchmarks with more speed levels use them on few tasks and to study other factors: VLALagBench~\cite{wen2026vlalagbench} uses nine speeds on four tasks to study inference latency, and AHEAD~\cite{syed2026ahead} uses four speeds in one conveyor scenario to evaluate its predictive world model.
In contrast, \proposed measures how the success rate itself changes with speed, at 13 belt speeds across 80 tasks in 16 skill categories (\cref{tab:benchmarks}).

\paragraph{Visual history and motion representations.}
In behavior cloning, observation history can lower performance because the policy learns to copy its own past actions from it~\cite{dehaan2019causal,wen2020copycat,chuang2022copycat}, and recent VLA methods add dedicated history mechanisms~\cite{liu2026ttfvla,koo2025hamlet,lin2026hifvla}.
MoveBench~\cite{zhang2026ppc} argues that VLAs trained on single frames are structurally blind to temporal dynamics.
For moving objects, DynamicVLA~\cite{xie2026dynamicvla} succeeds more often with an earlier frame, whereas DOMINO finds little benefit from raw-frame history on a 10-task subset and instead combines optical flow with future-state prediction. DynaWM uses observation history in a world model that regenerates VLA action chunks.
Motion can also be given to a policy explicitly, as frame differences~\cite{wang2016tsn}, optical flow~\cite{simonyan2014twostream}, or point traces~\cite{zheng2025tracevla}.
Another route to motion is video pretraining~\cite{wu2024gr1,hu2025vpp}: Cosmos Policy~\cite{kim2026cosmospolicy} and Fast-WAM~\cite{fastwam2026} adapt pretrained video generation models for control, and Fast-WAM retains video prediction during training without generating future frames at test time.
These studies examine history within their own methods.
We instead compare VLA and world-action models using the current frame alone or the current frame plus two past frames, and measure the effect at each belt speed.
As DynamicVLA does, we feed past frames through each model's own image input, without past actions, so \proposed measures what standard fine-tuning extracts from visible motion.

\paragraph{Probing and speed conditioning.}
Linear probes test whether a variable can be decoded from intermediate features~\cite{alain2017probe}.
However, decodability alone does not show that a model uses that information~\cite{elazar2021amnesic,hermann2020texture,fu2025hidden}, so we relate speed probes to changes in success rate caused by visual history and explicit speed input.
Conditioning a policy on physical parameters estimated from recent observations has precedents in online system identification~\cite{yu2017uposi} and rapid motor adaptation~\cite{kumar2021rma}, and DynamicWAM~\cite{lou2026dynamicwam} conditions a world-action model on image-plane velocity from optical flow.
Our image-based speed adapter follows this pattern and measures how much of the gain from ground-truth speed remains when speed must be estimated.

\section{\proposed}
\label{sec:benchmark}

\proposed evaluates manipulation success rates on 80 conveyor-belt tasks in 16 skill categories at 13 constant belt speeds (0.10--0.40~m/s).
We use ManiSkill~3~\cite{tao2024maniskill3} with a Franka Panda arm beside the conveyor.
Further benchmark details are provided in the supplementary material.

\subsection{Task Suites and Demonstrations}
\label{sec:tasks}
\paragraph{Task design.}
\cref{fig:tasks} shows one representative task from each of the 16 skill categories.
Categories group tasks by the main action their instruction asks for, inspired by DROID~\cite{khazatsky2024droid}, which measures task diversity by instruction verbs.
Each task has a language instruction, a simulator-defined success condition, and a scripted expert that uses the object's pose and velocity to generate demonstrations.
To ensure that each task is solvable at the tested speeds, we include a task only if its expert succeeds in at least 80\% of evaluation episodes at each of four representative speeds: 0.10, 0.20, 0.30, and 0.40~m/s.

\paragraph{Static counterpart.}
Every task also has a static counterpart with the belt stopped and the object initially within reach.
We treat it as a separate task that checks whether a policy can learn the manipulation with a stationary object, since its initial object pose and success condition may differ.

\paragraph{Environment.}
Demonstrations and evaluation episodes draw from 185 backgrounds, including indoor scenes from ReplicaCAD~\cite{szot2021habitat2}, ArchitecTHOR~\cite{deitke2022procthor}, and RoboCasa~\cite{nasiriany2024robocasa}, as well as a plain table.
We also randomize the platform material (39 options), lighting, the object (668 across all task sets), belt direction, the entry point (within 0.10~m), and exterior camera positions (up to 5~cm per axis).
The policy receives a $224 \times 224$ image tiling three exterior cameras and a $224 \times 224$ wrist image at 20~Hz.
Episodes last at most 15~s.

\paragraph{Demonstrations.}
For each of the 80 tasks, we collect 200 successful scripted-expert demonstrations at belt speeds drawn uniformly from 0.10--0.40~m/s (20 in each of ten equal-width bins; 16,000 in total).
Like a person picking from a conveyor, the expert reaches for an object as soon as it comes into view, so it grasps slow objects earlier on the belt than fast ones.
For 40 of the tasks, we also collect 200 demonstrations each on an accelerating belt (\cref{sec:benchmark_design}), with an acceleration drawn uniformly from $-0.05$ to $+0.05$~m/s$^2$ per episode (8,000 in total).

\subsection{Controlled Belt Speed and Visual Motion Cues}
\label{sec:benchmark_design}
In the main evaluation, the belt runs at a constant speed $v$ during each episode.
The randomized factors of \cref{sec:tasks}, such as the background, do not depend on $v$, and for each task and episode index we reuse the same draws at all 13 speeds.
The belt moves objects through friction, so they can slip or turn when the robot touches them, and a lifted object no longer moves with the belt.
The belt texture scrolls at speed $v$, so the speed is visible on the belt itself as well as in the object's motion.
A faster belt brings the object to the robot sooner and keeps it in view for a shorter time; we treat both as part of the effect of speed.
In the acceleration setting, the belt speed changes at a constant rate $a$ while the object is within 0.5~m of the robot, reaching 0.25~m/s when the object passes the robot.

\subsection{Speed-Wise Evaluation Protocol}
\label{sec:eval_protocol}
The main simulation comparison 
evaluates policies on all 80 tasks at 13 belt speeds from 0.10 to 0.40~m/s in 0.025~m/s increments.
Each evaluation episode uses the same randomized scene for every policy.
Simulation pauses during policy inference so that the success rate reflects each policy's actions rather than its inference latency~\cite{black2025rtc}, which depends on model size and hardware.
The acceleration evaluation uses the 40 accelerating-belt tasks at $a \in \{-0.05, -0.025, 0, +0.025, +0.05\}$~m/s$^2$, 20 episodes per task, paired across accelerations.

For simulated evaluations, let $S_v$ denote the mean success rate across the evaluated tasks at speed $v$, with each task weighted equally.
For the main 80-task comparison, the overall success rate is the mean of $S_v$ across all 13 speeds.
To test whether a speed-wise curve peaks inside the range, we fit least-squares lines of $S_v$ against $v$ separately to the speeds below 0.25~m/s and to those above it (six each), giving slopes $\beta_{\mathrm{slow}}$ and $\beta_{\mathrm{fast}}$ in points per 0.1~m/s, and define the peak score $P = \min(\beta_{\mathrm{slow}}, -\beta_{\mathrm{fast}})$.
$P > 0$ means that the success rate rises on the slow side and falls on the fast side. We say a policy's success rate peaks at intermediate speeds when the 95\% interval of $P$ lies above zero.
We compute 95\% confidence intervals for $S_v$, policy differences, and $P$ by resampling tasks.

\begin{figure*}[t]
  \centering
  \includegraphics[width=\textwidth]{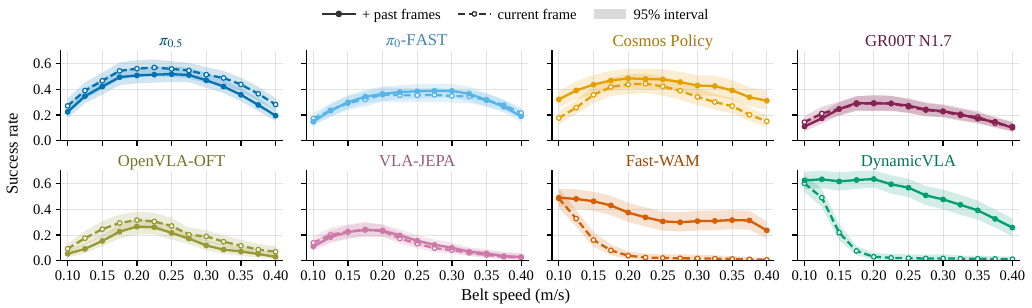}
\caption{\textbf{Success rate versus belt speed.}
Eight models fine-tuned with the current frame alone (dashed)
or with two past frames (solid).
All 80 tasks; 1,600 episodes per point;
bands: 95\% intervals from resampling tasks.}
  \label{fig:speed}
\end{figure*}

\section{Experiments}
Training and real-robot details and further results are provided in the supplementary material.
\subsection{Setup}
\label{sec:setup}

\paragraph{Models.}
The six VLA models are $\pi_{0.5}$~\cite{intelligence2025pi05}, $\pi_0$-FAST~\cite{pertsch2025fast}, OpenVLA-OFT~\cite{kim2025oft}, GR00T~N1.7~\cite{nvidia2025groot}, VLA-JEPA~\cite{vlajepa2026}, and DynamicVLA~\cite{xie2026dynamicvla}.
DynamicVLA is designed and pretrained for manipulating moving objects.
The two world-action models build on pretrained video diffusion models: Cosmos Policy~\cite{kim2026cosmospolicy} predicts actions within Cosmos-Predict2~\cite{nvidia2025cosmospredict2}, whereas Fast-WAM~\cite{fastwam2026} adds an action module to Wan2.2~\cite{wan2025wan}.

\paragraph{Training.}
We build on each model's released code (openpi~\cite{openpi} for $\pi_{0.5}$ and $\pi_0$-FAST) and extend it to take past frames.
We keep each model's recipe and change mainly the batch size and the number of training steps.
The speed-band, speed-conditioned, and static-counterpart policies are each fine-tuned on all 80 tasks together and evaluated on a 20-task subset.
We drew these tasks at random before those experiments, stratified by skill category and by the current-frame $\pi_{0.5}$ policy's success rate, split into three equal groups, so that the subset spans the difficulty levels of the full set and most of its skill categories.
The acceleration experiments use the 40 tasks with accelerating-belt demonstrations, a separate set.
For acceleration, we also fine-tune $\pi_{0.5}$, Fast-WAM, VLA-JEPA, and DynamicVLA with past frames on the accelerating-belt demonstrations, starting from the same pretrained weights as the constant-speed policies.

\paragraph{Visual history and control.}
For each model, we fine-tune a current-frame and a past-frame policy on the same demonstrations with the same recipe.
Both receive current exterior and wrist images; the past-frame policy also receives images from 0.1 and 0.2~s earlier.
Past frames enter through each model's native input format.
Six models use joint-position inputs and targets, while Fast-WAM and DynamicVLA use end-effector poses and targets tracked by inverse kinematics.
At 20~Hz, policies execute eight actions per prediction (0.4~s), except VLA-JEPA, which executes seven (0.35~s).

\paragraph{Speed conditioning and static counterparts.}
We fine-tune $\pi_{0.5}$, Fast-WAM, VLA-JEPA, and DynamicVLA with past frames and ground-truth belt speed.
The speed is normalized to $[-1,1]$. $\pi_{0.5}$ uses a sinusoidal embedding, while the other three add zero-initialized state inputs.
We compare their success rates with those of past-frame policies trained under the same recipe without speed input.
We separately fine-tune $\pi_{0.5}$ and Fast-WAM on static counterparts and use their success rates to check whether policies can learn the tasks without motion.

\subsection{Speed-Wise Policy Evaluation}
\label{sec:results}

\begin{table}[t]
\centering
\caption{\textbf{Success rate (\%) across 13 speeds.} \emph{Current}: current frame only; \emph{+Past}: two past frames added; \emph{Gain}: +Past minus Current, in percentage points. Brackets: 95\% bootstrap CIs over tasks.}
\label{tab:categories}
\scriptsize
\setlength{\tabcolsep}{3pt}
\begin{tabular*}{\columnwidth}{@{\extracolsep{\fill}}lrrrrr@{}}
\toprule
 & \multicolumn{2}{c}{Mean (\%)} & & \multicolumn{2}{c}{$P$ (points per 0.1\,m/s)} \\
\cmidrule(lr){2-3}\cmidrule(l){5-6}
Model & Current & +Past & Gain & Current & +Past \\
\midrule
$\pi_{0.5}$ & 46.1 & 40.5 & $-$5.6 & +20.8\,{\tiny[+17.1,+24.2]} & +22.9\,{\tiny[+18.6,+26.8]} \\
$\pi_0$-FAST & 30.4 & 31.3 & +0.9 & +10.8\,{\tiny[+7.7,+13.9]} & +16.1\,{\tiny[+13.0,+18.6]} \\
OpenVLA-OFT & 19.4 & 13.9 & $-$5.5 & +11.4\,{\tiny[+8.5,+14.5]} & +10.6\,{\tiny[+7.9,+13.4]} \\
GR00T N1.7 & 21.8 & 21.3 & $-$0.5 & +10.6\,{\tiny[+7.6,+12.9]} & +11.4\,{\tiny[+8.3,+14.6]} \\
Cosmos Policy & 32.1 & 41.6 & +9.6 & +18.6\,{\tiny[+14.4,+23.0]} & +11.8\,{\tiny[+7.3,+15.0]} \\
VLA-JEPA & 13.0 & 13.6 & +0.6 & +3.2\,{\tiny[+0.3,+5.9]} & +6.8\,{\tiny[+4.2,+9.0]} \\
DynamicVLA & 12.0 & 51.5 & +39.5 & $-$50.4\,{\tiny[$-$56.8,$-$44.0]} & $-$1.5\,{\tiny[$-$4.8,+1.7]} \\
Fast-WAM & 9.5 & 35.9 & +26.4 & $-$37.0\,{\tiny[$-$43.2,$-$30.8]} & $-$12.7\,{\tiny[$-$17.3,$-$8.3]} \\
\bottomrule
\end{tabular*}
\end{table}

We use \proposed to characterize speed-dependent failures and test whether visual history or explicit speed information improves the success rate.

\begin{figure}[t]
  \centering
  \includegraphics[width=\linewidth]{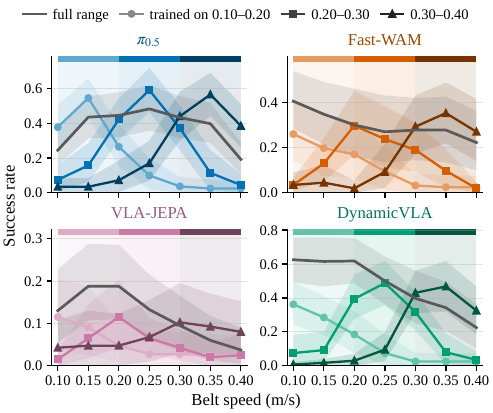}
\caption{\textbf{Success rate versus belt speed for policies trained on one third of the speed range.}
Top bars: training bands; grey: full-range training.
20-task subset, 400 episodes per point; all with past frames; bands: 95\% intervals from resampling tasks.
Panel y-scales differ.}  
  \label{fig:bands}
\end{figure}

\begin{figure}[t]
  \centering
  \includegraphics[width=\linewidth]{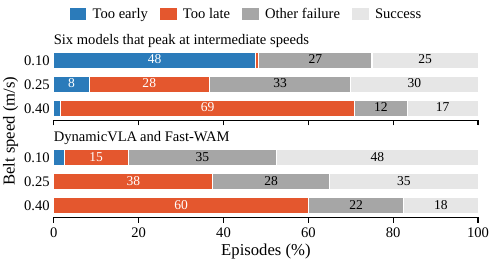}
  \caption{\textbf{Hand-labeled outcomes at three belt speeds.}
    Past-frame policies, one episode per subset task.
    Top: the six models that peak at intermediate speeds (120 episodes per speed); bottom: DynamicVLA and Fast-WAM (40).}
  \label{fig:qual-failures}
\end{figure}

\paragraph{How does the success rate vary with belt speed?}
\label{sec:curves}
For six of the eight models, the success rate peaks at intermediate speeds.
Their success rate rises from 0.10~m/s to a maximum and falls again by 0.40~m/s, with both current-frame and past-frame input (\cref{fig:speed}).
For all 12 of these policies, the peak score $P$ ranges from $+3.2$ to $+22.9$ points per 0.1~m/s, each with a 95\% interval above zero (\cref{tab:categories}).
The other two, DynamicVLA and Fast-WAM, do not peak and succeed most at or near the slow end of the range.
With past frames, $\pi_{0.5}$ and Fast-WAM have similar mean success rates (40.5\% and 35.9\%), so averaging over speeds hides this difference.

The 12 peaked policies fail even at slow speeds, although a slower belt leaves more time to act.
Slow speeds are solvable, as DynamicVLA and Fast-WAM succeed at 0.10~m/s in 48--62\% of episodes, more often than any of the 12 policies.
As a qualitative failure analysis, we labeled by hand the failures of the eight past-frame policies in one episode of each subset task at 0.10, 0.25, and 0.40~m/s (\cref{fig:qual-failures}).
For the six models that peak, 48\% of all episodes at 0.10~m/s fail because the policy acts before the object arrives, and 69\% at 0.40~m/s because it acts after the object has passed.
In other words, these policies tend to produce actions suited to the average demonstration speed rather than to the actual one.
Since the expert grasps slow objects earlier on the belt, a policy that ignores the speed learns the average timing, which fits mid-range speeds.

The peak indeed moves with the training speeds.
We fine-tune $\pi_{0.5}$, Fast-WAM, VLA-JEPA, and DynamicVLA separately on each of three speed bands, using past frames.
All 12 resulting policies succeed most often within their training band and less often than their full-range counterparts outside it (\cref{fig:bands}).
At 0.40~m/s, $\pi_{0.5}$ trained on 0.30--0.40~m/s reaches a 39\% success rate, compared with 19\% for the full-range policy, despite using one third as much training data.

\paragraph{Does visual history improve the success rate?}
\label{sec:history}
We compare current-frame and past-frame policies trained on the same demonstrations spanning the full speed range.
Past frames raise the mean success rate by 26.4, 9.6, and 39.5 percentage points for Fast-WAM, Cosmos Policy, and DynamicVLA, respectively (\cref{tab:categories}).
The other five models change by $-5.6$ to $+0.9$ points.
For Fast-WAM and DynamicVLA, the gain extends to the highest speed: at 0.40~m/s, they reach 24\% and 26\% success rates with history, versus near zero without it (\cref{fig:speed}).
Past frames also change the shape of the curves, which the mean gain does not show.
For Fast-WAM, Cosmos Policy, and DynamicVLA, $P$ moves toward zero: they gain little at their best speed and more at the other speeds, so past frames widen the range of speeds they handle.
Unlike the other five models, these three were pretrained on multiple frames, in different ways.
Fast-WAM and Cosmos Policy build on video diffusion models (Wan2.2 and Cosmos-Predict2) trained to generate the frames of a video together, whereas DynamicVLA was trained to predict actions from temporally separated images of moving objects~\cite{fastwam2026,kim2026cosmospolicy,xie2026dynamicvla}.
VLA-JEPA also learns from videos, but during pretraining its policy pathway receives only an initial observation; future-frame features serve as prediction targets~\cite{vlajepa2026}.
This distinction is consistent with VLA-JEPA's small history gain: predicting future features alone does not guarantee a benefit from observed visual history.

\begin{figure}[t]
  \centering
  \includegraphics[width=\linewidth]{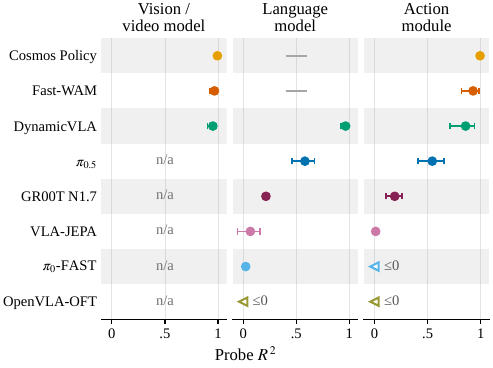}
\caption{\textbf{Linear decodability of belt speed by policy stage.}
Probe $R^2$ on held-out tasks with 95\% CIs; 12 tasks of the 20-task subset at all 13 speeds, with the arm held at its start pose while the object passes.
Triangles: $R^2 \leq 0$; n/a: the vision encoder encodes each image separately, so the current image's tokens cannot see the past frames; dashes: no such stage; the action module of $\pi_0$-FAST and Cosmos Policy: the action tokens of their language and video model.}
  \label{fig:probe}
\end{figure}

\paragraph{Does speed decodability explain history gains?}
\label{sec:probe}
To examine whether history gains are reflected in policy representations, we fit linear probes for belt speed at three stages of each fine-tuned policy: the vision or video model, the language model, and the action module.
For Fast-WAM, Cosmos Policy, and DynamicVLA, which gain substantially from past frames, speed is strongly decodable at the action module: $R^2=0.93$ for Fast-WAM, $0.99$ for Cosmos Policy, and $0.86$ for DynamicVLA (\cref{fig:probe}).
The speed is less linearly decodable from the other five models: $\pi_{0.5}$ reaches $R^2 = 0.58$ in its language model, and $\pi_0$-FAST, OpenVLA-OFT, GR00T N1.7, and VLA-JEPA at most 0.21.
None of these five gains more than one point from past frames (\cref{tab:categories}), which suggests that past frames help little when the speed is hard to decode from the policy.
Even before fine-tuning, the speed is far more decodable from the pretrained models of Cosmos Policy (Cosmos-Predict2, $R^2 = 0.84$), Fast-WAM (Wan2.2, $0.57$), and DynamicVLA ($0.77$) than from those of the other five (at most 0.08).
These results suggest that pretraining on multiple frames, from video~\cite{hu2025vpp} or from moving-object demonstrations, helps a policy act on the speed.

\begin{figure}[t]
  \centering
  \includegraphics[width=\linewidth]{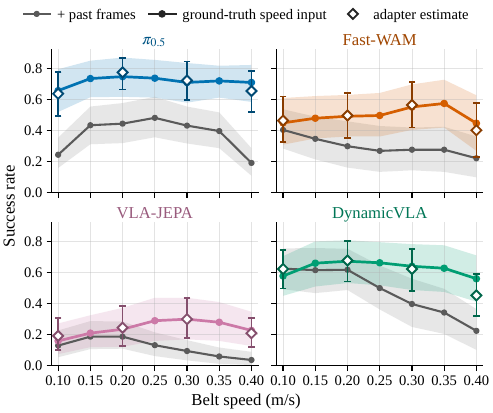}
\caption{\textbf{Success rate with ground-truth or estimated belt speed.}
20-task subset, all with past frames; bands and bars: 95\% intervals from resampling tasks.}
  \label{fig:told}
\end{figure}

\paragraph{Belt-speed input helps, even when estimated.}
\label{sec:told}
\label{sec:adapter}
Belt-speed input improves all four models tested, and a speed estimated from the images keeps most of this gain.
On the 20-task subset at seven speeds (0.10--0.40~m/s in steps of 0.05), ground-truth speed input raises the mean success rate of the past-frame policies by 34, 20, 12, and 15 percentage points for $\pi_{0.5}$, Fast-WAM, VLA-JEPA, and DynamicVLA (\cref{fig:told}).
$\pi_{0.5}$ and VLA-JEPA gain from the speed even though past frames do not raise their success rate.
We then give the same policies, without retraining, a speed estimated from the images instead.
The speed adapter, a small head on frozen RAFT-small~\cite{teed2020raft}, estimates the belt speed from two exterior frames 0.1~s apart, with a mean absolute error of 0.010~m/s on held-out episodes.
Before the first estimate exists, the policy plans with a prior of 0.25~m/s and re-plans once the estimate is available (control step 2).
Averaged over 0.10, 0.20, 0.30, and 0.40~m/s, the estimate recovers 98\%, 98\%, 105\%, and 83\% of the gain from the ground-truth speed for $\pi_{0.5}$, Fast-WAM, VLA-JEPA, and DynamicVLA, respectively.

\begin{figure}[t]
  \centering
  \includegraphics[width=\linewidth]{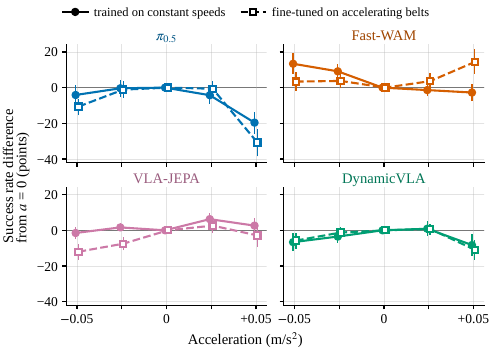}
  \caption{\textbf{Change in success rate with belt acceleration.}
    Each curve is a policy's success rate minus its own success rate at zero acceleration on paired episodes.
    One panel per model; solid: the past-frame policy trained on constant speeds; dashed: the same model fine-tuned on accelerating belts.
    40 tasks $\times$ 20 episodes per point; bars: 95\% paired bootstrap CIs over tasks.}
  \label{fig:accel}
\end{figure}

\paragraph{Can policies learn to handle an accelerating belt?}
\label{sec:accel}
To test whether policies can learn to handle a belt speed that changes within an episode, we fine-tune $\pi_{0.5}$, Fast-WAM, VLA-JEPA, and DynamicVLA on accelerating belts and compare each with its constant-speed policy on these 40 tasks (\cref{fig:accel}).
When the belt speeds up ($a=+0.05$~m/s$^2$), Fast-WAM's accelerating-belt policy gains 14 points over its success rate at $a=0$, whereas its constant-speed policy changes by $-3$.
The accelerating-belt policies of the other three models do not gain, changing by $-3$ to $-31$ points.
Fast-WAM and DynamicVLA both gain from past frames at constant speeds, yet only Fast-WAM's accelerating-belt policy gains when the belt speeds up.
Even after acceleration training, DynamicVLA's success falls by 11 points at $a=+0.05$~m/s$^2$.
Gaining from past frames at constant speeds therefore does not ensure that a policy adapts to a speed that changes within an episode.

\subsection{Real-Robot Evaluation}
\label{sec:real}
We test whether the simulated effects of past frames and speed input hold on a real robot.

\begin{figure}[t]
  \centering
  \includegraphics[width=\columnwidth]{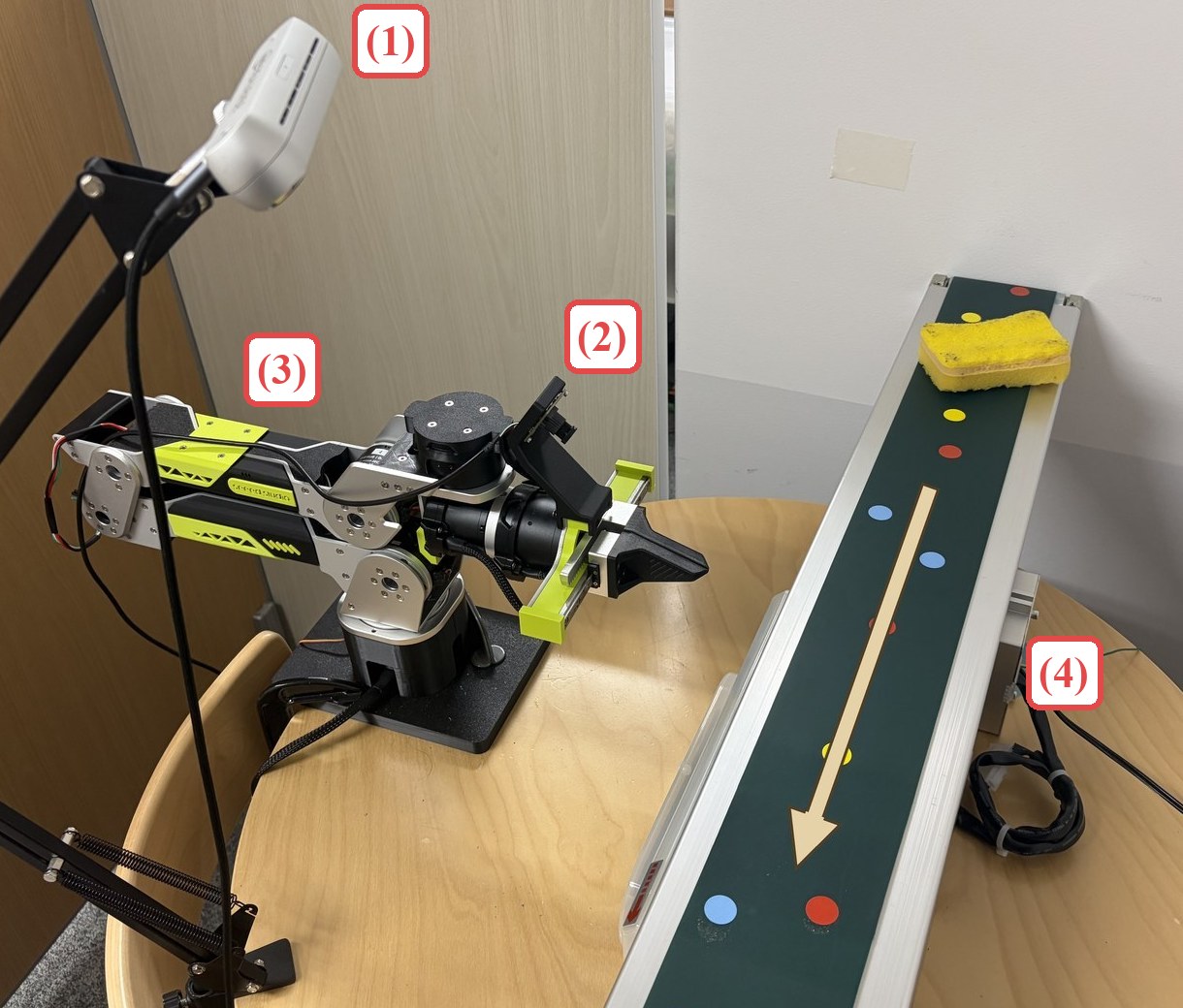}
  \caption{\textbf{Real-robot setup.}
  (1) Fixed head camera (Intel RealSense D455f), (2) wrist camera (Innomaker U20CAM), (3) reBot arm, and (4) 1~m conveyor.
  Colored stickers make belt motion visible.
  Arrow: belt direction.
  Both camera images are provided at $224 \times 224$, and the arm is controlled at 20~Hz.
  The image shows the start of a sponge-picking trial.}
  \label{fig:real-setup}
\end{figure}

\paragraph{Setup.}
A reBot arm operates beside a 1~m conveyor with a fixed head camera and a wrist camera (\cref{fig:real-setup}).
We use three tasks, ``Pick the yellow sponge'', ``Place the red cube in the plate'', and ``Push the apple off the belt''.
We collect 120 demonstrations, with 10 per task at each of four speeds from 0.05 to 0.20~m/s, and fine-tune $\pi_{0.5}$ and Fast-WAM on 108 of them, keeping 12 for validation.
Each model is fine-tuned with the current frame and with two past frames.
We also fine-tune $\pi_{0.5}$ with past frames and the belt speed set for each demonstration, as in \cref{sec:told}, and run it with the speed set for the trial and with the adapter's estimate.
We run 10 trials per task and speed in each of these six configurations, yielding 120 trials each and 720 in total.
A completed task scores 1, target contact scores 0.5, and a failure without contact scores 0.
Inference runs on a remote H100 GPU while the arm continues moving, using the latency-aware action streaming introduced with DynamicVLA~\cite{xie2026dynamicvla}. Median request-to-response time ranges from 63 to 118~ms.

\begin{figure}[t]
  \centering
  \includegraphics[width=\columnwidth]{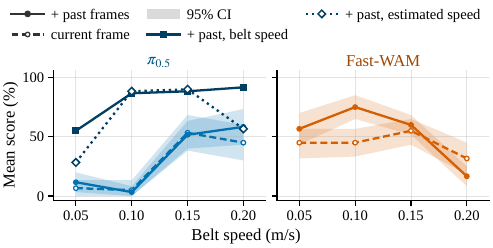}
  \caption{\textbf{Real-robot score vs.\ belt speed}, averaged over the three tasks (30 trials per point).
  Dark lines: $\pi_{0.5}$ with past frames and the belt speed, set for the trial (squares) or estimated by the adapter (diamonds).
  Bands: 95\% bootstrap intervals over trials, for the policies without speed input.}
  \label{fig:real-results}
\end{figure}

\paragraph{Results.}
Past frames raise Fast-WAM's mean score from 44 to 52 and $\pi_{0.5}$'s from 28 to 31, but neither difference is significant ($p=0.11$ and $0.46$; \cref{fig:real-results}).
At the speeds shared with simulation (0.10--0.20~m/s), past frames add 7 and 3 points for Fast-WAM and $\pi_{0.5}$, against 21 and $-5$ in simulation.

Belt-speed input helps on the robot as well: $\pi_{0.5}$ with past frames scores 80 with the ground-truth speed and 66 with the adapter's estimate, against 31 without speed input (both $p<0.001$).

\begin{figure}[t]
  \centering
  \includegraphics[width=\columnwidth]{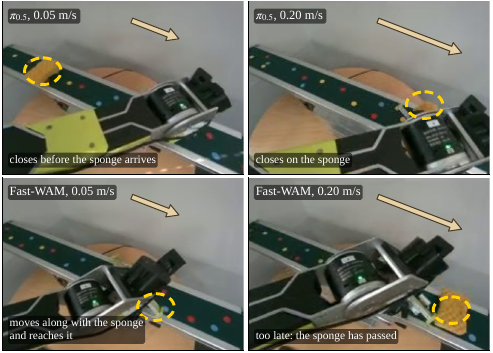}
  \caption{\textbf{Gripper position during real-conveyor picking.}
  One trial per policy and speed, both policies using past frames, shown from the head camera when the gripper first closes.
  Each trial is closest to the median closing time among its 10 trials; Fast-WAM never closes at 0.20~m/s, so its last frame is shown.
  Dashed outline: sponge.
  Arrow: belt direction, shorter at the slower speed.}
  \label{fig:real-qualitative}
\end{figure}

Qualitatively, in picking (\cref{fig:real-qualitative}), $\pi_{0.5}$ closes its gripper at about the same place at 0.05 and 0.20~m/s, too early for the slower sponge, whereas Fast-WAM follows the sponge at 0.05~m/s but arrives too late at 0.20~m/s.

\section{Conclusion}
\label{sec:conclusion}

We introduced \proposed, a benchmark of 80 conveyor tasks at 13 belt speeds.
Many fine-tuned policies tend to produce actions suited to the average demonstration speed, and our results suggest that past frames widen the speed range only for those pretrained on multiple frames.
An estimated speed recovers most of the gain that policies obtain from the ground-truth speed.

\paragraph{Limitations.}
Due to computational constraints, each policy is fine-tuned once, which may introduce some noise into our results.
The eight models differ in architecture and training data, so we cannot attribute the gain from past frames to pretraining on multiple frames alone.
The simulator pauses during inference, so inference latency does not affect the success rates.
Other motion inputs, such as optical flow, might help policies that gain little from raw frames.

\section*{Acknowledgments}
This work used the computational resources of Wisteria/BDEC-01 at the Information Technology Center, The University of Tokyo; Miyabi at the Joint Center for Advanced High Performance Computing (JCAHPC); Genkai at the Research Institute for Information Technology, Kyushu University; and the TSUBAME4.0 supercomputer at Institute of Science Tokyo.
Claude Code, with Anthropic's Claude models, wrote most of the code, ran and monitored the experiments, and drafted and edited the text under our direction, and Claude Code, OpenAI Codex, and Google Gemini gave feedback on our drafts.
We designed the study, checked all code, results, and text, and take full responsibility for the paper.

\maketitlesupplementary
\setcounter{figure}{0}
\setcounter{table}{0}
\renewcommand{\thefigure}{S\arabic{figure}}
\renewcommand{\thetable}{S\arabic{table}}
\appendix
\crefalias{section}{appendix}\crefalias{subsection}{subappendix}\crefalias{subsubsection}{subsubappendix}
\raggedbottom
\renewcommand{\topfraction}{0.9}\renewcommand{\bottomfraction}{0.8}\renewcommand{\textfraction}{0.07}
\renewcommand{\floatpagefraction}{0.8}\renewcommand{\dbltopfraction}{0.9}\renewcommand{\dblfloatpagefraction}{0.8}
\setcounter{topnumber}{4}\setcounter{bottomnumber}{2}\setcounter{totalnumber}{6}\setcounter{dbltopnumber}{3}

\makeatletter
\setlength{\@fptop}{0pt}
\setlength{\@fpsep}{\floatsep}
\setlength{\@fpbot}{0pt plus 1fil}
\setlength{\@dblfptop}{0pt}
\setlength{\@dblfpsep}{\dblfloatsep}
\setlength{\@dblfpbot}{0pt plus 1fil}
\makeatother

\noindent This supplement gives the details behind each part of the main paper.
\begin{itemize}
  \item \cref{sec:supp-bench}: the tasks, randomized factors, belt model, demonstrations, static counterparts, and accelerating belt (\cref{sec:benchmark}).
  \item \cref{sec:supp-policies}: the inputs, execution, and fine-tuning of each policy (\cref{sec:setup}).
  \item \cref{sec:supp-protocol}: the evaluation protocol and statistics (\cref{sec:eval_protocol}).
  \item \cref{sec:supp-results}: full results and controls for \cref{sec:results}, including the static counterparts and the accelerating belt.
  \item \cref{sec:supp-probe}: the probes (\cref{sec:probe}).
  \item \cref{sec:supp-adapter}: the speed-estimation adapter (\cref{sec:adapter}).
  \item \cref{sec:supp-real}: the real-robot study (\cref{sec:real}).
  \item \cref{sec:supp-repro}: licenses and release.
\end{itemize}

\section{Benchmark}
\label{sec:supp-bench}

\subsection{Tasks and success conditions}
\label{sec:supp-tasks}
\begin{figure*}[p]
  \centering
  \includegraphics[width=\textwidth]{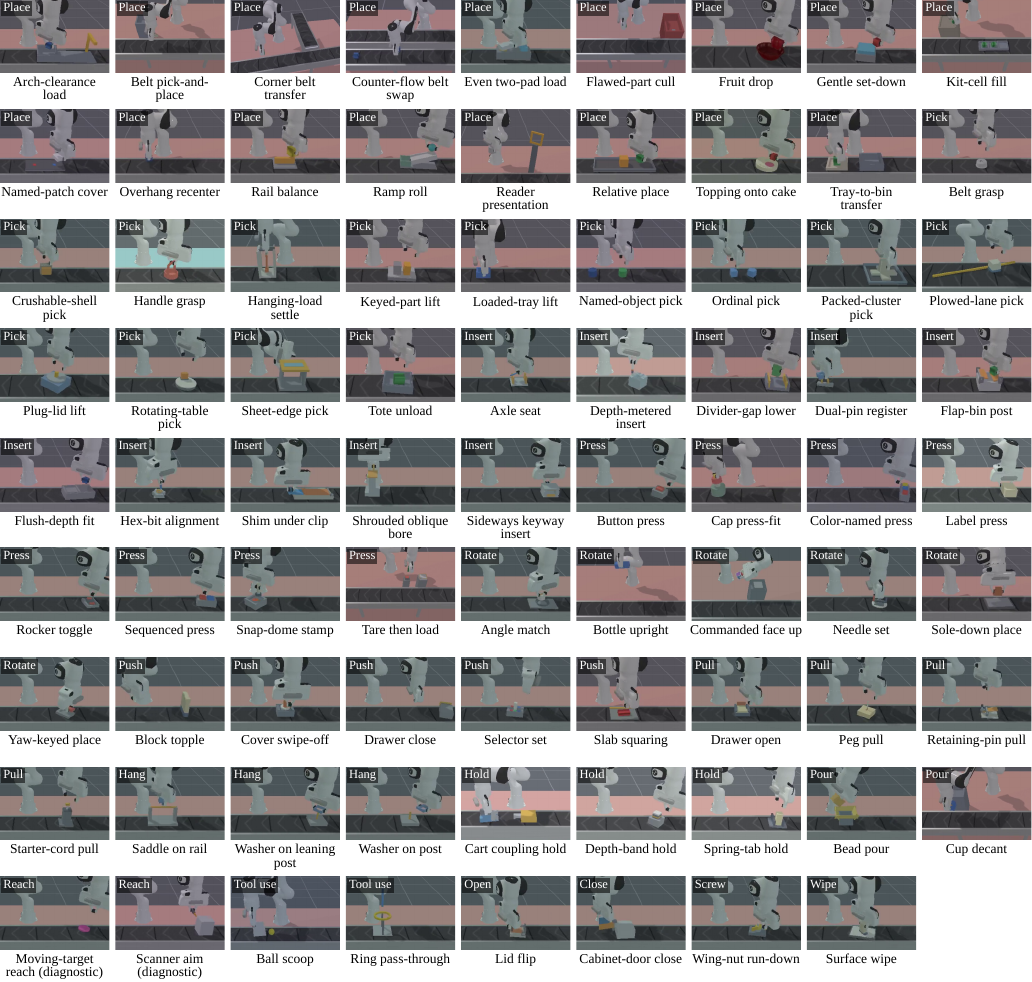}
  \caption{\textbf{All 80 tasks}, grouped by skill category, one frame each from a scripted-expert demonstration, rendered without the randomized table and background.}
  \label{fig:supp-alltasks}
\end{figure*}

\cref{tab:supp-tasks} lists the 80 tasks with their instructions and the scripted expert's success rates, the released code gives their success conditions, and \cref{fig:supp-alltasks} shows one frame of each.
The scripted expert succeeds in at least 80\% of episodes of every task at each of 0.10, 0.20, 0.30, and 0.40\,m/s (lowest rate 82.3\%), over 384 separate check episodes per task and speed.

\paragraph{Task subsets.}
We drew the 20-task subset ($\dagger$ in \cref{tab:supp-tasks}) at random before any experiment on it, stratified by skill category and by three equal-sized groups of the current-frame $\pi_{0.5}$ policy's success rate.
The next 20 tasks of the same random draw ($\ddagger$) complete the 40-task subset, which is used for the end-effector comparison (\cref{sec:supp-ee}).
The acceleration experiments (\cref{sec:accel}) use a different set of 40 tasks, the accelerating-belt tasks ($\S$; \cref{sec:supp-capabilities}).

\subsection{Randomized factors}
\label{sec:supp-random}
An episode's random draws are the same at every speed (\cref{sec:benchmark_design}).
\cref{tab:supp-random} lists the factors shared by all tasks. Most tasks also randomize factors of their own, such as the object's yaw or its offset across the belt; the released code gives each task's ranges.
\begin{table*}[t]
\centering
\caption{\textbf{Factors randomized in every task.} Training and evaluation draw each factor from the same distribution, except that evaluation fixes the belt speed; $\mathcal{U}[a, b]$ is uniform.}
\label{tab:supp-random}
\small
\begin{tabular}{llp{11.0cm}}
\toprule
Factor & Type & Range \\
\midrule
Belt speed $v$ & continuous & training: 0.10--0.40\,m/s, 20 demonstrations per task in each of ten equal-width bins, uniform within a bin; evaluation: 13 values 0.100, 0.125, \dots, 0.400\,m/s \\
Belt direction & discrete (2) & $\{+y, -y\}$ along the belt (the object moves to the robot's left or right), equal probability; the object enters from the upstream end \\
Entry point & continuous & moved downstream from the task's start point by $\mathcal{U}[0.00, 0.10]$\,m \\
Background & discrete (185) & ReplicaCAD 90, RoboCasa 84, ArchitecTHOR 10, plain table 1 \\
Platform material & discrete (39) & physically based rendering (PBR) materials for the table the conveyor stands on \\
Exterior camera position & continuous & position offset $\mathcal{U}[-5, 5]$\,cm per axis, drawn separately for each of the 3 exterior cameras (head camera, $+y$ side camera, and $-y$ side camera); look-at point fixed; wrist camera not perturbed \\
Lighting & continuous & intensity scale $\mathcal{U}[0.7, 1.3]$; color scale per channel $\mathcal{U}[0.85, 1.15]$; key-light direction offset $\mathcal{U}[-0.4, 0.4]$ per component \\
Clutter & discrete (5) & 0--4 visual-only cubes (no collision) on the far side of the belt, half edge length $\mathcal{U}[1.5, 3.5]$\,cm, random hue \\
Object instance & discrete (1--602) & from the task's object set: 917 task--object pairs; 49 tasks use one fixed object, and the other 31 draw from 668 distinct objects; the demonstrations contain 391 of the pairs, so evaluation also draws task--object pairs never seen in training \\
\bottomrule
\end{tabular}
\end{table*}

\paragraph{Scripted expert.}
The scripted expert holds the arm still until an arriving object first appears in an exterior camera, then moves faster when the belt is faster or the object is about to leave the belt, up to ten times its base limit of 1\,cm per control step.
To vary the motion between demonstrations, a slowly varying random factor scales this limit on each axis to 70--100\%; above 0.20\,m/s the lower end rises with the belt speed, to 80\% at 0.40\,m/s.
Waiting for the object and moving faster on a faster belt make the expert's timing, and so the length and shape of the demonstrations, depend on the speed (\cref{tab:supp-demos}).
In 6 tasks, the expert also draws its own aim point or approach parameter in each episode, which changes the demonstrations but not the scene.

\paragraph{Objects of Belt pick-and-place.}
We keep an object for Belt pick-and-place only if the scripted expert succeeds with it in at least 80\% of episodes at each of 0.10, 0.30, and 0.50\,m/s, which cover the 0.10--0.40\,m/s range of the benchmark; 602 of the 1,547 candidate objects pass.
Each object gets 32 episodes per speed; with 22--29 successes at a speed, it gets 64 more episodes there and needs at least 77 successes in the 96.

\subsection{Belt model}
\label{sec:supp-belt}
\paragraph{Belt drive.}
The conveyor is a fixed box, so the simulator never moves the belt surface.
Instead, before every physics step ($\Delta t = 0.01$\,s), we apply a friction force to each object in contact with the belt.
While static friction can hold the object (coefficient 0.6; 1.0 in Plowed-lane pick), this force keeps it moving at the belt speed; otherwise kinetic friction (coefficient 0.4; 0.8 in Plowed-lane pick) opposes the slip.
An object on a fixture that rides the belt, such as a tray, moves with the fixture.

\paragraph{Belt surface.}
The belt texture repeats every 0.2\,m and scrolls at the belt speed in every camera, and it is the same in every episode.
The belt is 1.8\,m long and 16\,cm wide, with its surface 25\,cm above the table.

\paragraph{Start point.}
Each task's start point lies 0.40--0.86\,m upstream of the robot base (median 0.81\,m), so at the median the object reaches the robot after about 8\,s at 0.10\,m/s and 2\,s at 0.40\,m/s.

\subsection{Demonstrations}
\label{sec:supp-demos}
\begin{figure}[t]
  \centering
  \includegraphics[width=\linewidth]{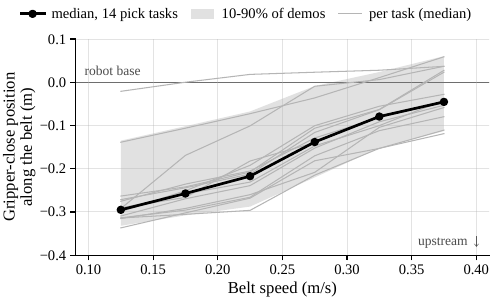}
  \caption{\textbf{Grasp position of the scripted expert versus belt speed.}
  Position along the belt where the expert first closes the gripper, against the belt speed (0 is the robot base, negative is upstream).
  All 2,800 demonstrations of the 14 pick tasks, in speed bins of 0.05\,m/s; black line: median over these demonstrations; thin lines: median per task; band: 10th to 90th percentile over demonstrations.}
  \label{fig:supp-expert-grasp}
\end{figure}

Each task has 200 successful scripted-expert demonstrations, 20 in each of ten equal-width speed bins.
\cref{tab:supp-demos} summarizes them in 0.05\,m/s bins.
Evaluation episodes, unlike the demonstrations, are not filtered by the expert's success.
Slow demonstrations are longer and so contribute more frames, and none of the fine-tuning recipes reweights frames by speed (\cref{sec:supp-peak} tests weighting every demonstration equally).
The expert reaches for an object as soon as it comes into view, so it grasps slow objects earlier on the belt than fast ones (\cref{fig:supp-expert-grasp}).
An expert that waited at a fixed point would make slow speeds easy even for a policy that ignores the speed; with this expert, a policy must adapt to the speed to succeed at both slow and fast speeds.

\begin{table}[t]
\centering
\caption{\textbf{Training demonstrations in 0.05\,m/s belt-speed bins.} Share of the 16,000 demonstrations and of their 1,259,153 frames, and length in control steps (mean $\pm$ s.d.).}
\label{tab:supp-demos}
\footnotesize
\begin{tabular}{lrrr}
\toprule
Speed (m/s) & Demonstrations (\%) & Frames (\%) & Length (steps) \\
\midrule
0.10--0.15 & 16.2 & 22.5 & 109.2 $\pm$ 32.1 \\
0.15--0.20 & 16.9 & 18.8 & 87.6 $\pm$ 28.5 \\
0.20--0.25 & 16.9 & 16.3 & 76.3 $\pm$ 27.3 \\
0.25--0.30 & 16.5 & 15.0 & 71.5 $\pm$ 28.5 \\
0.30--0.35 & 16.8 & 14.1 & 66.1 $\pm$ 28.1 \\
0.35--0.40 & 16.7 & 13.2 & 62.2 $\pm$ 27.1 \\
\midrule
All & 100 & 100 & 78.7 $\pm$ 32.7 \\
\bottomrule
\end{tabular}
\end{table}

\subsection{Static counterparts}
\label{sec:supp-static}
Every task has a static counterpart, with the belt stopped and the object within reach.
Success conditions that refer to the moving belt, such as keeping an object moving with it, are dropped or replaced.
In eight tasks the static counterpart differs in more than the start pose and these conditions, and in two more one word of the instruction also differs (\cref{tab:supp-static}).
\begin{table}[t]
\centering
\caption{\textbf{Static counterparts that differ from their moving task} beyond the start pose and the belt-related success conditions.}
\label{tab:supp-static}
\footnotesize
\begin{tabular}{@{}p{2.4cm}p{5.4cm}@{}}
\toprule
Task & What differs in the static counterpart \tabularnewline
\midrule
\raggedright Arch-clearance load & \raggedright the arch does not move past the gripper, so the arm must withdraw behind it \tabularnewline
\raggedright Bead pour & \raggedright one word of the instruction \tabularnewline
\raggedright Cart coupling hold & \raggedright the robot slides the front cart back onto the rear cart instead of holding it while the belt brings the rear cart \tabularnewline
\raggedright Corner belt transfer & \raggedright stricter check that the other object on the belt stays undisturbed \tabularnewline
\raggedright Counter-flow belt swap & \raggedright stricter check that the other object on the belt stays undisturbed \tabularnewline
\raggedright Flawed-part cull & \raggedright a defect-free part must stay where it is instead of riding past \tabularnewline
\raggedright Ordinal pick & \raggedright blocks are counted from the belt's downstream end instead of by arrival; stricter check on the other block \tabularnewline
\raggedright Plowed-lane pick & \raggedright the block starts where the plow would have pushed it \tabularnewline
\raggedright Relative place & \raggedright ``in front of'' refers to a fixed side of the tray \tabularnewline
\raggedright Surface wipe & \raggedright one word of the instruction \tabularnewline
\bottomrule
\end{tabular}
\end{table}

\cref{sec:supp-static-results} tests whether policies can learn the static counterparts.

\subsection{Accelerating belt}
\label{sec:supp-capabilities}
On the accelerating belt, the speed depends on $u$, the position the object would have if it had stayed on the belt, measured along the travel direction ($u < 0$ upstream of the robot):
$v(u)^2 = v_c^2 + 2a\,\mathrm{clip}(u, -L/2, L/2)$, with speed $v_c$ at the robot, acceleration $a$, and $L = 1$\,m the length of the stretch, centered on the robot, over which the speed changes.
Because $u$ follows the belt travel since the start, holding or lifting an object does not change the belt speed.
We keep $v_c = 0.25$\,m/s and vary $a$ within $\pm0.05$\,m/s$^2$, so the speed stays within 0.11--0.34\,m/s, inside the constant-speed training range.

\paragraph{Tasks and demonstrations.}
The accelerating belt uses 40 of the 80 tasks ($\S$ in \cref{tab:supp-tasks}).
They are chosen by skill category from the 68 tasks in which the scripted expert succeeds in at least 29 of 32 episodes at each of $a = -0.05$, 0, and $+0.05$\,m/s$^2$.
Each task has 200 demonstrations with $a$ drawn uniformly from $[-0.05, 0.05]$\,m/s$^2$ per episode.

\FloatBarrier
\section{Policies}
\label{sec:supp-policies}

\subsection{Inputs and execution}
\label{sec:supp-inputs}
\cref{tab:supp-inputs} lists what each model receives and outputs, and how the past frames and the speed input enter it.
\cref{alg:supp-exec} gives the execution loop of every simulated evaluation; the adapter runs also re-plan once at step 2 (\cref{sec:supp-adapter}).
\providecommand{\suppL}{\let\suppcr\\\raggedright\let\\\suppcr\hspace{0pt}}
\providecommand{\suppNote}[1]{\par\hangindent=0.7em\hangafter=1\noindent\makebox[0.7em][l]{\textsuperscript{#1}}}
\begin{table*}[t]
\centering
\caption{\textbf{Inputs and outputs of each model} at control step $t$ (20\,Hz). Every model receives a $224 \times 224$ exterior image (the two side-camera views on top, the head-camera view below) and a $224 \times 224$ wrist image. The past-frame policies add the exterior images from $t-4$ and $t-2$ (0.2 and 0.1\,s earlier; DynamicVLA also the wrist images). EE: end-effector; IK: inverse kinematics.}
\label{tab:supp-inputs}
\scriptsize
\setlength{\tabcolsep}{2pt}
\renewcommand{\arraystretch}{1.15}
\begin{tabular}{@{}p{0.085\textwidth}*{8}{p{0.105\textwidth}}@{}}
\toprule
& \suppL $\pi_{0.5}$
& \suppL $\pi_0$-FAST
& \suppL OpenVLA-OFT
& \suppL GR00T N1.7
& \suppL Cosmos Policy
& \suppL VLA-JEPA
& \suppL DynamicVLA
& \suppL Fast-WAM \\
\midrule
\suppL Pretrained weights
& \suppL $\pi_{0.5}$ base
& \suppL $\pi_0$-FAST base
& \suppL OpenVLA 7B
& \suppL GR00T N1.7 3B
& \suppL Cosmos-Predict2 2B Video2World
& \suppL VLA-JEPA pretraining checkpoint
& \suppL DynamicVLA pretrained on DOM
& \suppL Wan2.2-TI2V-5B with the authors' action expert \\
\suppL How past frames enter
& \suppL 2 extra image slots
& \suppL 2 extra image slots
& \suppL 2 extra images
& \suppL 2 extra images
& \suppL 2 extra latent frames
& \suppL 2 extra images
& \suppL stacked as input channels; past wrist images too
& \suppL 2 extra latent frames \\
\suppL Robot state
& \suppL joints + gripper
& \suppL joints + gripper
& \suppL joints + gripper
& \suppL joints + gripper
& \suppL joints + gripper
& \suppL joints + gripper
& \suppL EE pose
& \suppL EE pose \\
\suppL Action
& \suppL joint targets + gripper
& \suppL joint targets + gripper
& \suppL joint targets + gripper
& \suppL joint targets + gripper
& \suppL joint targets + gripper
& \suppL joint targets + gripper
& \suppL EE pose + gripper (IK)
& \suppL EE pose + gripper (IK) \\
\suppL Action chunk / executed steps
& \suppL 16 / 8
& \suppL 16 / 8
& \suppL 16 / 8
& \suppL 16 / 8
& \suppL 16 / 8
& \suppL 7 / 7
& \suppL 20 / 8
& \suppL 32 / 8 \\
\suppL Speed input
& \suppL embedding added to the action expert's conditioning
& \suppL --
& \suppL --
& \suppL --
& \suppL --
& \suppL extra state input
& \suppL extra state input
& \suppL extra state input \\
\bottomrule
\end{tabular}
\end{table*}

\begin{figure}[t]
\caption{\textbf{Execution of one evaluation episode.} The loop and constants used for every policy in all our simulated evaluations.}
\label{alg:supp-exec}
\footnotesize
\noindent\rule{\linewidth}{0.8pt}
\begin{tabbing}
\hspace{1.1em}\=\hspace{1.1em}\=\hspace{1.1em}\=\kill
\textbf{Input:} policy $\pi$, task, belt speed $v$, episode index $k$\\
\textbf{Constants:} 20\,Hz control, 5 physics steps of 10\,ms each;\\
\>maximum episode length $T = 300$ steps (15\,s); $H = 8$ executed steps\\
\>per chunk (VLA-JEPA: 7); past-frame offsets $\mathcal{D} = \{-4, -2\}$\\
\>(current-frame policies: $\mathcal{D} = \emptyset$)\\[3pt]
$s \gets$ a fixed seed of the task \` $\triangleright$ same at every speed\\
draw the randomized scene from $(s, k)$; set the belt speed to $v$\\
\textbf{for} $t = 0, 1, \dots, T-1$ \textbf{do}\\
\>observe $o_t$: exterior image, wrist image, robot state\\
\>\textbf{if} $t = 0$ \textbf{or} $t - t_0 = \min(H, |A|)$ \textbf{then}\\
\>\>$A \gets \pi\bigl(\{o_{t+d}\}_{d \in \mathcal{D}}, o_t, \text{instruction}\,[, v']\bigr)$\\
\>\>$t_0 \gets t$ \` $\triangleright$ the simulation waits for $\pi$\\
\>$a \gets A[t - t_0]$\\
\>\textbf{if} $\pi$ outputs joint targets \textbf{then}\\
\>\>clip the 7 arm targets to the joint limits\\
\>\textbf{else}\\
\>\>solve IK for the absolute EE pose from the current joints\\
\>run the PD controller and physics for 50\,ms\\
\>\textbf{if} the task's success condition holds \textbf{then return} success\\
\>\textbf{if} the task can no longer succeed \textbf{then return} failure\\
\textbf{return} failure
\end{tabbing}
\vspace{-8pt}
\noindent\rule{\linewidth}{0.8pt}\par\vspace{2pt}
{\footnotesize\noindent
A past observation $o_{t+d}$ contains only the exterior image (DynamicVLA: also the wrist image); frames before the episode start repeat the first frame (Fast-WAM masks them instead).
$v' = (v - 0.25)/0.15$ is given only to the speed-conditioned policies.
An episode can no longer succeed when, for example, the object has passed out of reach downstream.
\par}
\end{figure}

\subsection{Fine-tuning recipes}
\label{sec:supp-recipes}
Every model starts from public pretrained weights and is fine-tuned with its authors' code, except VLA-JEPA, for which we wrote a training loop around the released model.
\cref{tab:supp-recipes} gives the main settings, and \cref{tab:supp-deviations} lists where we depart from the authors' configuration and why.
\providecommand{\suppL}{\let\suppcr\\\raggedright\let\\\suppcr\hspace{0pt}}
\providecommand{\suppNote}[1]{\par\hangindent=0.7em\hangafter=1\noindent\makebox[0.7em][l]{\textsuperscript{#1}}}
\begin{table*}[t]
\centering
\caption{\textbf{Fine-tuning recipes.} The main settings for fine-tuning each model on the 16,000 demonstrations (DynamicVLA: 15,840; \cref{tab:supp-deviations}), and $\pi_{0.5}$ on the 108 real-robot ones; \cref{tab:supp-deviations} lists where they differ from the authors' configuration, and the released code gives the full configurations.}
\label{tab:supp-recipes}
\scriptsize
\setlength{\tabcolsep}{1.5pt}
\renewcommand{\arraystretch}{1.15}
\begin{tabular}{@{}p{0.07\textwidth}*{9}{p{0.095\textwidth}}@{}}
\toprule
& \suppL $\pi_{0.5}$ (simulation)
& \suppL $\pi_{0.5}$ (real robot)
& \suppL $\pi_0$-FAST
& \suppL OpenVLA-OFT
& \suppL GR00T N1.7
& \suppL Cosmos Policy
& \suppL VLA-JEPA
& \suppL DynamicVLA
& \suppL Fast-WAM \\
\midrule
\suppL Trained parts
& \suppL all
& \suppL ${=}$
& \suppL all
& \suppL LoRA (rank 32, $\alpha = 16$) on every linear layer; action head and state projector
& \suppL action head, state and action encoders; VLM frozen
& \suppL diffusion transformer
& \suppL all but the V-JEPA~2 encoder
& \suppL all
& \suppL video and action transformers, state projector \\
\suppL Trained / total parameters
& \suppL 3.35B / 3.35B
& \suppL ${=}$
& \suppL 2.92B / 2.92B
& \suppL 296M / 7.84B
& \suppL 1.62B / 3.14B
& \suppL 1.96B / 2.08B
& \suppL 2.44B / 2.77B
& \suppL 430M / 430M
& \suppL 6.02B / 6.72B \\
\suppL Peak learning rate
& \suppL $5{\times}10^{-5}$
& \suppL $2.5{\times}10^{-5}$
& \suppL $2.5{\times}10^{-5}$
& \suppL $5{\times}10^{-4}$
& \suppL $10^{-4}$
& \suppL $10^{-4}$
& \suppL $10^{-5}$ VLM, $10^{-4}$ action head, $3{\times}10^{-5}$ rest
& \suppL $10^{-4}$
& \suppL $10^{-4}$ \\
\suppL Batch size
& \suppL 256
& \suppL 32
& \suppL 32
& \suppL 64
& \suppL 128
& \suppL 128
& \suppL 64
& \suppL 32
& \suppL 128 \\
\suppL Steps
& \suppL 30,000
& \suppL 1,125
& \suppL 130,000
& \suppL 25,000
& \suppL 90,000
& \suppL 15,838
& \suppL 45,000
& \suppL 155,872
& \suppL 9,838 \\
\suppL Epochs
& \suppL 6.10
& \suppL 2.37
& \suppL 3.30
& \suppL 1.27
& \suppL 9.15
& \suppL 1.61
& \suppL 2.29
& \suppL 3.96
& \suppL 1.00 \\
\bottomrule
\end{tabular}
\par\vspace{3pt}
\begin{minipage}{\textwidth}
\scriptsize\raggedright
${=}$: same as the column to its left. Epochs count passes over the 1.26M frames of the 16,000 demonstrations (real robot: the 15,204 frames of its 108 demonstrations).
\end{minipage}
\end{table*}

\providecommand{\suppL}{\let\suppcr\\\raggedright\let\\\suppcr\hspace{0pt}}
\begin{table*}[t]
\centering
\caption{\textbf{Deviations from the authors' configuration}, with the reason. The authors' training lengths are set for their own datasets, so we set ours for our 16,000 demonstrations within our compute budget. A dash in the last column: no further reason.}
\label{tab:supp-deviations}
\scriptsize
\setlength{\tabcolsep}{3pt}
\renewcommand{\arraystretch}{1.15}
\begin{tabular}{@{}p{0.125\textwidth}p{0.1\textwidth}p{0.22\textwidth}p{0.19\textwidth}p{0.285\textwidth}@{}}
\toprule
\suppL Model & \suppL Setting & \suppL Authors' value & \suppL Ours & \suppL Reason or remark \\
\midrule
\suppL $\pi_{0.5}$ (simulation) & \suppL Learning rate, schedule, warm-up, batch, EMA & \suppL openpi's default configuration & \suppL openpi's LIBERO configuration & \suppL The real-robot $\pi_{0.5}$ keeps openpi's default configuration. \\
\suppL $\pi_0$-FAST & \suppL Steps & \suppL openpi's default configuration: 30,000 steps & \suppL 130,000 steps & \suppL -- \\
\suppL OpenVLA-OFT & \suppL Steps & \suppL LIBERO configuration: 150,005 steps & \suppL 25,000 steps & \suppL -- \\
& \suppL Action chunk & \suppL 8 (LIBERO configuration) & \suppL 16 & \suppL The same chunk as $\pi_{0.5}$, $\pi_0$-FAST, GR00T N1.7, and Cosmos Policy, which execute 8 of 16 predicted steps. \\
\suppL GR00T N1.7 & \suppL Batch, steps & \suppL default configuration: batch 64, 10,000 steps; LIBERO configuration: batch 640, 20,000 steps & \suppL batch 128, 90,000 steps & \suppL -- \\
\suppL Cosmos Policy & \suppL Batch, data, length & \suppL LIBERO configuration: batch 1,920, policy rollouts for half of each batch; 40,000 steps & \suppL 15,838 steps (1.61 epochs) at batch 128 on demonstrations only & \suppL We have no policy rollouts, so it learns from the demonstrations alone, as the other models do. \\
\suppL VLA-JEPA & \suppL Batch, steps & \suppL batch 256, 30,000 steps & \suppL batch 64 (16 samples $\times$ 4 gradient-accumulation steps); 45,000 steps & \suppL -- \\
& \suppL Training code & \suppL the authors' training script & \suppL our own training loop & \suppL Re-implements the authors' training script with the same model, parameter groups, optimizer, and loss. \\
\suppL DynamicVLA & \suppL Trained parts & \suppL released configuration: vision encoder, connector, and language model frozen & \suppL all & \suppL The DynamicVLA paper trains the full model. \\
& \suppL Batch, length & \suppL released configuration: 500 epochs at batch 44 & \suppL 155,872 steps at batch 32 & \suppL About 4 epochs over 15,840 demonstrations; the other 160 (2 per task) are held out for a validation loss. \\
& \suppL Action targets & \suppL deltas & \suppL absolute end-effector targets & \suppL The released code normalizes the delta targets with statistics of the absolute actions; on our data this made the loss diverge. \\
\suppL Fast-WAM & \suppL Optimizer & \suppL AdamW & \suppL 8-bit AdamW & \suppL AdamW's optimizer states did not fit in GPU memory. \\
& \suppL Length & \suppL LIBERO configuration: 10 epochs & \suppL one epoch at batch 128 & \suppL -- \\
\bottomrule
\end{tabular}
\end{table*}

\paragraph{Training seeds.}
To measure the effect of the training seed, we fine-tune the current-frame and past-frame $\pi_{0.5}$ policies with two more random seeds each (changing the data order, the augmentation, and the flow-matching noise) and evaluate the six policies on the same episodes of the 20-task subset at seven speeds.
With every seed, past frames lower the success rate by 3.8 to 5.8 points.
For both the current-frame and the past-frame policy, neither new seed's success rate differs significantly from the original seed's (Cochran--Mantel--Haenszel test stratified by task and speed, all $p > 0.1$).

\paragraph{Speed-band policies.}
Each speed-band policy of \cref{fig:bands} uses the recipe of the model's full-range past-frame policy on the demonstrations of one band (0.10--0.20, 0.20--0.30, or 0.30--0.40\,m/s; about a third of the demonstrations each).
It trains for the same number of epochs, so for fewer steps, and its learning-rate schedule is shortened by the same factor.

\section{Evaluation Protocol}
\label{sec:supp-protocol}

\paragraph{Episodes.}
The main evaluation runs each policy on 20 episodes per task and speed (20,800 per policy).
Because an episode's random draws depend only on the task and the episode index (\cref{sec:benchmark_design}), comparisons between policies are paired, and so are comparisons between speeds.
Experiments on the 20-task subset reuse these episodes at seven speeds, 0.10 to 0.40\,m/s in steps of 0.05\,m/s (2,800 per policy, 13\% of the full evaluation), and this subset can also serve as a cheaper protocol.
The adapter experiments use four of these speeds (0.10, 0.20, 0.30, and 0.40\,m/s), and the probes use separately recorded episodes (\cref{sec:supp-probe}).

\paragraph{Metrics.}
A \emph{task--speed pair} is one task at one speed.
The \emph{gain from past frames} is the mean over speeds of $S_v^{\text{past}} - S_v^{\text{current}}$.
The \emph{share of the gain recovered} by the adapter is $(S^{\text{adapter}} - S^{\text{none}}) / (S^{\text{GT}} - S^{\text{none}})$, where each $S$ is the mean of $S_v$ over the four adapter speeds for the past-frame policy without speed input (none) or the speed-conditioned policy given the adapter's estimate or the ground-truth speed (GT).

\paragraph{Intervals and tests.}
A range in square brackets after a value is its 95\% confidence interval.
In simulation, for a success rate, a difference, a share, or the peak score $P$, we resample tasks with replacement (2,000 bootstrap samples, or 10,000 for the accelerating belt), keeping all speeds and all compared policies of a task together, and take the 2.5th and 97.5th percentiles.
\cref{sec:supp-real} gives the intervals and tests on the real robot.
Each main policy is a single fine-tuning run ($\pi_{0.5}$ also has two more seeds; \cref{sec:supp-recipes}), so the intervals do not include the variation between training runs.
Intervals and $p$-values are given per comparison and are not corrected for multiple comparisons.

\section{Full Results}
\label{sec:supp-results}

\subsection{Success rate by speed and skill}
\begin{figure*}[p]
  \centering
  \includegraphics[width=\textwidth]{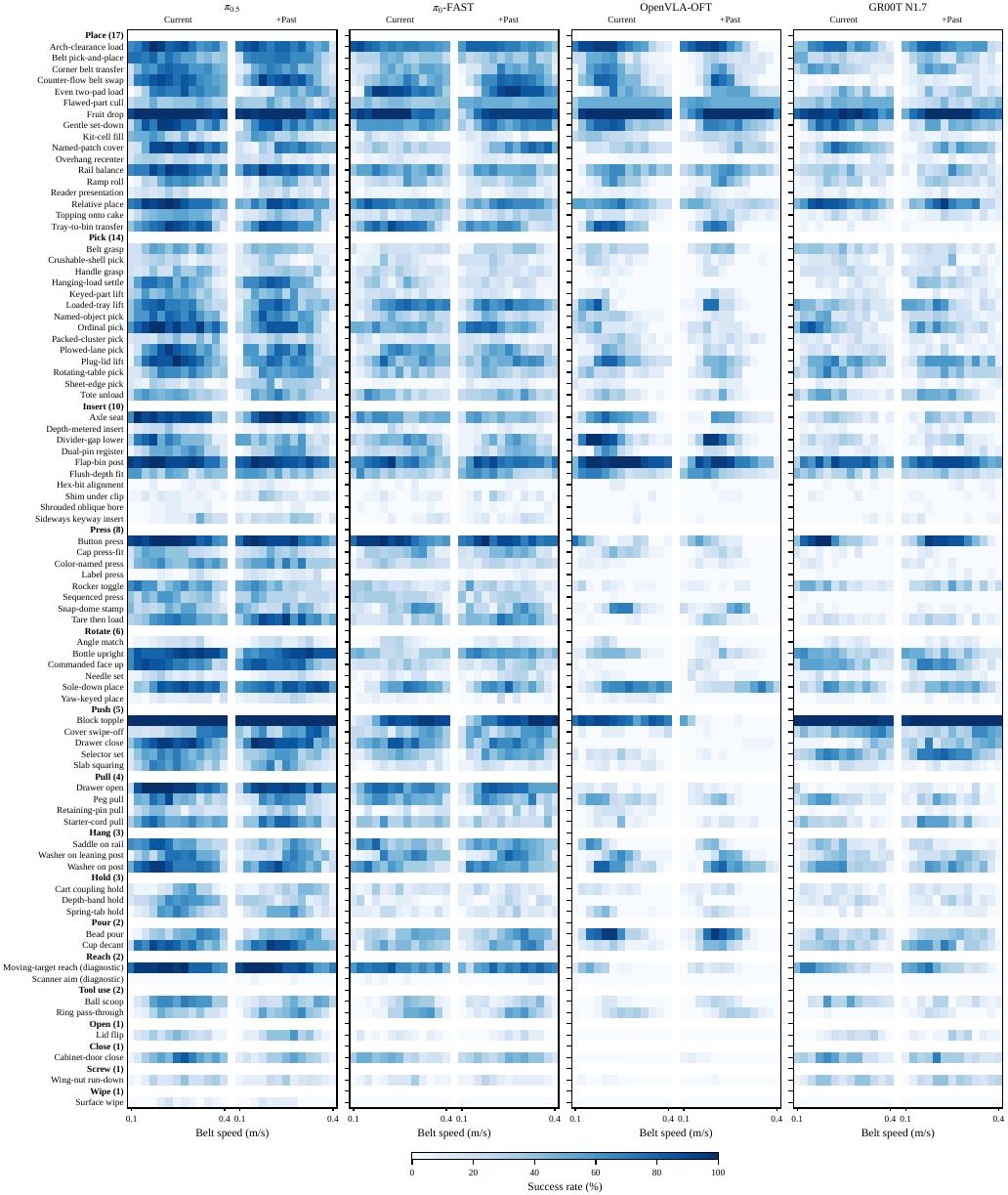}
  \caption{\textbf{Success rate of every task at every belt speed} for $\pi_{0.5}$, $\pi_0$-FAST, OpenVLA-OFT, and GR00T N1.7, with the current frame (left half of each panel) and with past frames (right half); 13 speeds from 0.10 to 0.40\,m/s, 20 episodes per cell.}
  \label{fig:supp-task-speed-a}
\end{figure*}

\begin{figure*}[p]
  \centering
  \includegraphics[width=\textwidth]{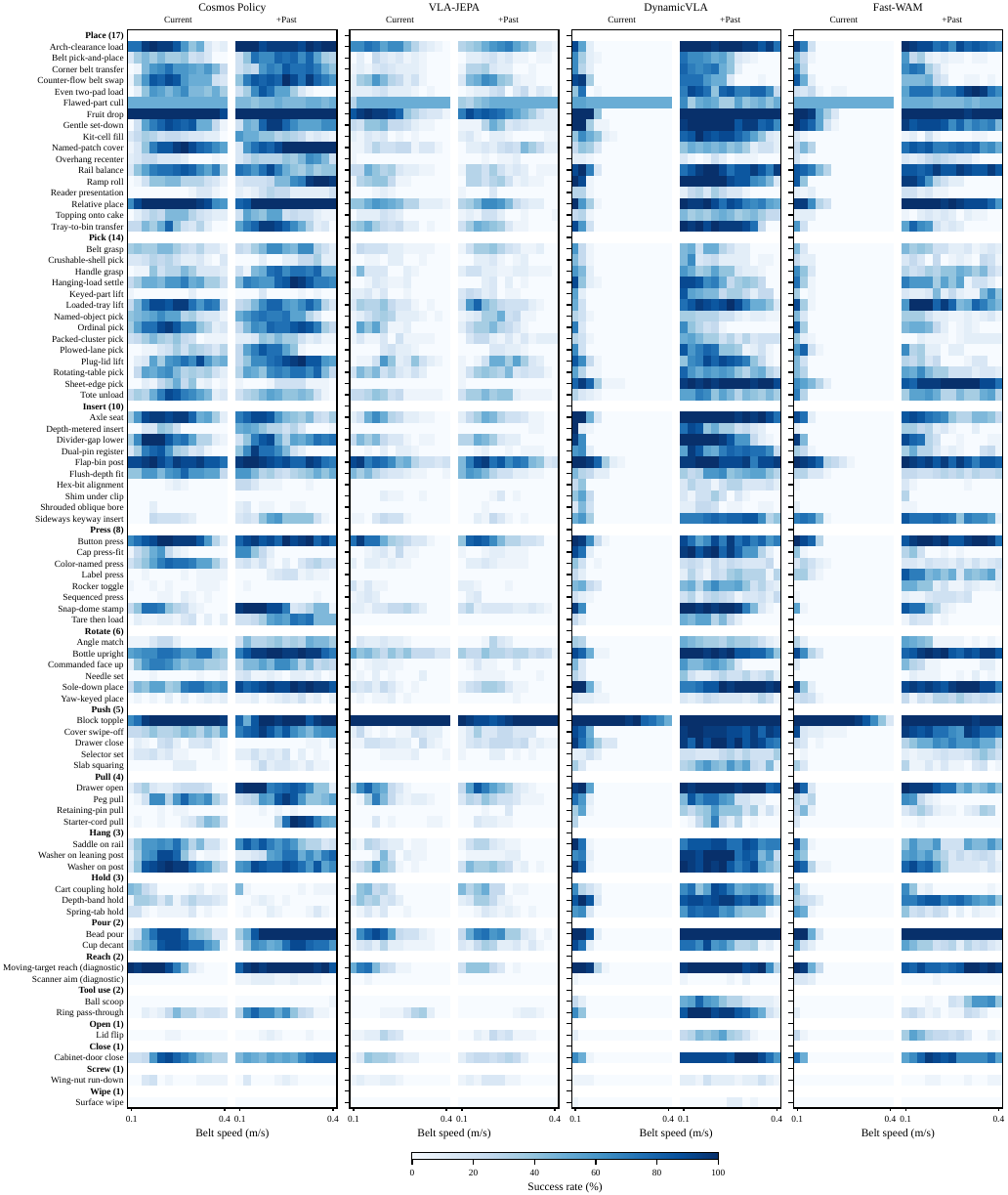}
  \caption{\textbf{Success rate of every task at every belt speed} for Cosmos Policy, VLA-JEPA, DynamicVLA, and Fast-WAM, as in \cref{fig:supp-task-speed-a}.}
  \label{fig:supp-task-speed-b}
\end{figure*}

\cref{tab:speed13} gives the success rate of every policy at every speed (the values in \cref{fig:speed}), \cref{tab:supp-categories} breaks \cref{tab:categories} down by skill category, and \cref{fig:supp-task-speed-a,fig:supp-task-speed-b} show every task at every speed.
\begin{table*}[t]
\centering
\caption{\textbf{Success rate (\%) at each belt speed} on all 80 tasks (20 episodes per task and speed). \emph{Cur.}: current frame only; \emph{+Past}: with two past frames. The 95\% intervals from resampling tasks (not shown) are about $\pm$5 points. Bold: highest rate in each row before rounding. Mean: over the 13 speeds.}
\label{tab:speed13}
\scriptsize
\setlength{\tabcolsep}{3pt}
\begin{tabular}{llrrrrrrrrrrrrrr}
\toprule
& & \multicolumn{13}{c}{Belt speed (m/s)} & \\
\cmidrule(lr){3-15}
Model & Input & 0.100 & 0.125 & 0.150 & 0.175 & 0.200 & 0.225 & 0.250 & 0.275 & 0.300 & 0.325 & 0.350 & 0.375 & 0.400 & Mean \\
\midrule
$\pi_{0.5}$ & Cur. & 27 & 39 & 47 & 54 & 56 & \textbf{57} & 56 & 55 & 51 & 49 & 44 & 36 & 28 & 46.1 \\
 & +Past & 22 & 35 & 42 & 49 & 51 & 51 & \textbf{52} & 51 & 47 & 42 & 36 & 28 & 20 & 40.5 \\
$\pi_0$-FAST & Cur. & 17 & 24 & 29 & 32 & \textbf{36} & 35 & 35 & 36 & 35 & 34 & 32 & 28 & 22 & 30.4 \\
 & +Past & 15 & 24 & 30 & 34 & 36 & 38 & 38 & \textbf{39} & 39 & 36 & 32 & 26 & 19 & 31.3 \\
OpenVLA-OFT & Cur. & 9 & 18 & 25 & 29 & \textbf{32} & 30 & 27 & 20 & 19 & 15 & 12 & 9 & 7 & 19.4 \\
 & +Past & 5 & 9 & 15 & 23 & \textbf{27} & 26 & 22 & 17 & 12 & 9 & 7 & 5 & 3 & 13.9 \\
GR00T N1.7 & Cur. & 14 & 21 & 25 & 28 & \textbf{29} & 29 & 26 & 24 & 23 & 20 & 18 & 14 & 11 & 21.8 \\
 & +Past & 11 & 17 & 25 & \textbf{29} & 29 & 29 & 27 & 24 & 23 & 21 & 17 & 15 & 10 & 21.3 \\
Cosmos Policy & Cur. & 18 & 26 & 36 & 42 & 44 & \textbf{44} & 42 & 39 & 34 & 30 & 27 & 20 & 15 & 32.1 \\
 & +Past & 32 & 39 & 44 & 47 & \textbf{48} & 48 & 48 & 46 & 43 & 42 & 39 & 34 & 31 & 41.6 \\
VLA-JEPA & Cur. & 14 & 20 & 23 & \textbf{24} & 23 & 17 & 13 & 10 & 8 & 6 & 4 & 3 & 3 & 13.0 \\
 & +Past & 11 & 18 & 22 & \textbf{24} & 24 & 20 & 16 & 12 & 10 & 7 & 6 & 4 & 3 & 13.6 \\
DynamicVLA & Cur. & \textbf{60} & 49 & 22 & 8 & 3 & 2 & 2 & 2 & 2 & 2 & 2 & 1 & 1 & 12.0 \\
 & +Past & 62 & 63 & 62 & 63 & \textbf{64} & 59 & 57 & 51 & 48 & 44 & 39 & 33 & 26 & 51.5 \\
Fast-WAM & Cur. & \textbf{48} & 33 & 16 & 8 & 4 & 3 & 2 & 2 & 2 & 2 & 1 & 1 & 1 & 9.5 \\
 & +Past & \textbf{49} & 48 & 46 & 43 & 37 & 34 & 31 & 30 & 31 & 31 & 32 & 31 & 24 & 35.9 \\
\bottomrule
\end{tabular}
\end{table*}

\begin{table*}[t]
\centering
\caption{\textbf{Success rate (\%) per skill category}, pooled over 13 belt speeds (0.10--0.40\,m/s) and 20 episodes per task and speed. \emph{Current}: current frame only; \emph{+Past}: with two past frames. \emph{Mean} and $P$: over all 80 tasks, as in \cref{tab:categories}; $P$ is the peak score (\cref{sec:eval_protocol}) in points per 0.1\,m/s, from the success rates at all 13 speeds, with a 95\% interval from resampling tasks in brackets.}
\label{tab:supp-categories}
\scriptsize
\setlength{\tabcolsep}{2.6pt}
\begin{tabular*}{\textwidth}{@{\extracolsep{\fill}}llrrrrrrrrrrrrrrrrrr@{}}
\toprule
 & & \multicolumn{16}{c}{Success rate (\%) per skill category (number of tasks)} & & \\
\cmidrule(lr){3-18}
Model & Input & \rotatebox{90}{Place (17)} & \rotatebox{90}{Pick (14)} & \rotatebox{90}{Insert (10)} & \rotatebox{90}{Press (8)} & \rotatebox{90}{Rotate (6)} & \rotatebox{90}{Push (5)} & \rotatebox{90}{Pull (4)} & \rotatebox{90}{Hang (3)} & \rotatebox{90}{Hold (3)} & \rotatebox{90}{Pour (2)} & \rotatebox{90}{Reach (2)} & \rotatebox{90}{Tool use (2)} & \rotatebox{90}{Open (1)} & \rotatebox{90}{Close (1)} & \rotatebox{90}{Screw (1)} & \rotatebox{90}{Wipe (1)} & \shortstack{Mean\\(\%)} & \shortstack{$P$ (points per 0.1\,m/s)\\{[95\% CI]}} \\
\midrule
$\pi_{0.5}$ & Current & 57 & 45 & 34 & 44 & 42 & 61 & 52 & 58 & 34 & 50 & 43 & 36 & 22 & 48 & 20 & 5 & 46 & +20.8\,{\tiny[+17.1,+24.2]} \\
 & +Past & 47 & 38 & 32 & 41 & 42 & 59 & 47 & 41 & 26 & 50 & 43 & 32 & 24 & 28 & 15 & 5 & 40 & +22.9\,{\tiny[+18.6,+26.8]} \\
\addlinespace[1pt]
$\pi_0$-FAST & Current & 42 & 27 & 24 & 30 & 20 & 41 & 38 & 41 & 12 & 35 & 34 & 22 & 6 & 33 & 10 & 0 & 30 & +10.8\,{\tiny[+7.7,+13.9]} \\
 & +Past & 42 & 27 & 24 & 33 & 23 & 44 & 35 & 45 & 13 & 38 & 32 & 22 & 7 & 35 & 10 & 0 & 31 & +16.1\,{\tiny[+13.0,+18.6]} \\
\addlinespace[1pt]
OpenVLA-OFT & Current & 35 & 15 & 24 & 9 & 15 & 21 & 13 & 22 & 6 & 26 & 7 & 9 & 0 & 0 & 1 & 0 & 19 & +11.4\,{\tiny[+8.5,+14.5]} \\
 & +Past & 29 & 10 & 16 & 8 & 9 & 2 & 5 & 21 & 6 & 26 & 4 & 11 & 0 & 2 & 0 & 0 & 14 & +10.6\,{\tiny[+7.9,+13.4]} \\
\addlinespace[1pt]
GR00T N1.7 & Current & 30 & 21 & 15 & 12 & 21 & 43 & 14 & 27 & 9 & 26 & 18 & 13 & 12 & 35 & 25 & 0 & 22 & +10.6\,{\tiny[+7.6,+12.9]} \\
 & +Past & 27 & 20 & 16 & 14 & 21 & 47 & 16 & 23 & 9 & 33 & 17 & 9 & 10 & 30 & 24 & 0 & 21 & +11.4\,{\tiny[+8.3,+14.6]} \\
\addlinespace[1pt]
Cosmos Policy & Current & 45 & 33 & 32 & 24 & 29 & 30 & 19 & 49 & 13 & 56 & 28 & 9 & 1 & 50 & 5 & 0 & 32 & +18.6\,{\tiny[+14.4,+23.0]} \\
 & +Past & 59 & 41 & 35 & 30 & 46 & 36 & 43 & 59 & 4 & 77 & 51 & 20 & 2 & 60 & 3 & 0 & 42 & +11.8\,{\tiny[+7.3,+15.0]} \\
\addlinespace[1pt]
VLA-JEPA & Current & 19 & 10 & 12 & 8 & 9 & 24 & 11 & 11 & 10 & 22 & 12 & 4 & 7 & 15 & 4 & 0 & 13 & +3.2\,{\tiny[+0.3,+5.9]} \\
 & +Past & 19 & 13 & 13 & 8 & 8 & 26 & 13 & 11 & 11 & 24 & 7 & 1 & 7 & 15 & 6 & 0 & 14 & +6.8\,{\tiny[+4.2,+9.0]} \\
\addlinespace[1pt]
DynamicVLA & Current & 15 & 8 & 13 & 9 & 9 & 27 & 10 & 13 & 14 & 18 & 13 & 5 & 1 & 9 & 1 & 0 & 12 & $-$50.4\,{\tiny[$-$56.8,$-$44.0]} \\
 & +Past & 58 & 41 & 51 & 42 & 46 & 69 & 46 & 80 & 64 & 73 & 46 & 55 & 25 & 88 & 10 & 2 & 52 & $-$1.5\,{\tiny[$-$4.8,+1.7]} \\
\addlinespace[1pt]
Fast-WAM & Current & 14 & 7 & 9 & 6 & 6 & 21 & 7 & 11 & 7 & 13 & 12 & 0 & 2 & 10 & 0 & 0 & 9 & $-$37.0\,{\tiny[$-$43.2,$-$30.8]} \\
 & +Past & 45 & 31 & 29 & 28 & 39 & 52 & 25 & 38 & 31 & 64 & 44 & 17 & 22 & 73 & 2 & 0 & 36 & $-$12.7\,{\tiny[$-$17.3,$-$8.3]} \\
\bottomrule
\end{tabular*}
\end{table*}

\begin{figure*}[t]
  \centering
  \includegraphics[width=\textwidth]{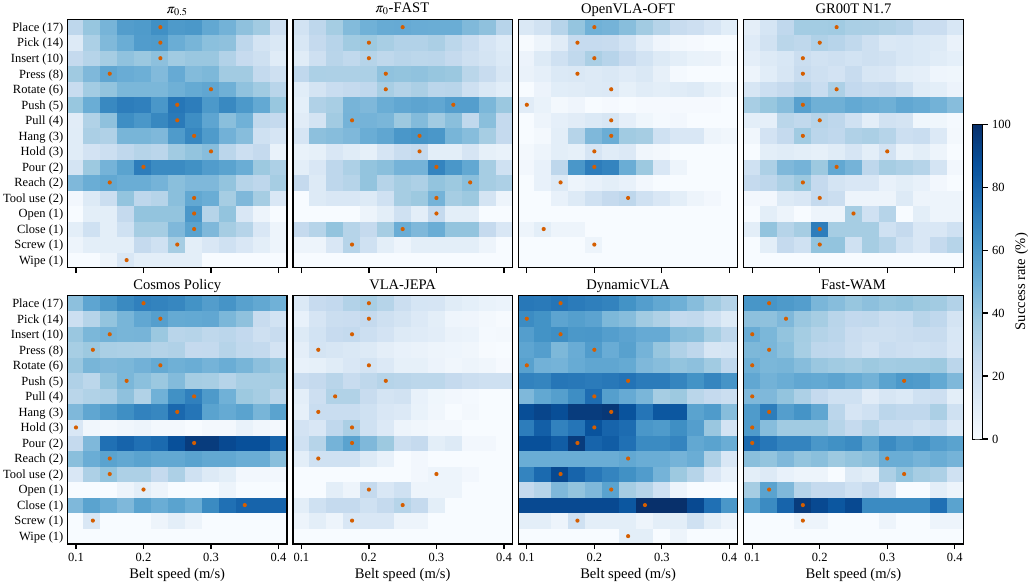}
  \caption{\textbf{Success rate per skill category and belt speed} for the eight policies with past frames (number of tasks in parentheses; each cell pools the category's tasks, 20 episodes per task and speed).
  Dots: the speed of highest success in each row; rows without a success have none.}
  \label{fig:supp-category-speed}
\end{figure*}

\paragraph{Skill categories and speed.}
Pooled over the 12 policies of the six models that peak at intermediate speeds, the success rate of every skill category is highest between 0.15 and 0.275\,m/s, and the 95\% interval of the peak score lies above zero in 10 of the 12 categories with two or more tasks (all but Push and Hold).
The peak therefore does not come from a few skills; \cref{fig:supp-category-speed} shows each past-frame policy separately.

\subsection{Is the peak at intermediate speeds an artifact?}
\label{sec:supp-peak}

\paragraph{Both sides of the peak.}
\cref{tab:supp-peak-slopes} gives the two slopes behind each policy's peak score.
For all 12 policies of the six models that peak, $\beta_{\mathrm{slow}} > 0$ and $\beta_{\mathrm{fast}} < 0$, with 95\% intervals that exclude zero.
\begin{table}[t]
\centering
\caption{\textbf{The two sides of the peak.} Slopes $\beta_{\mathrm{slow}}$ and $\beta_{\mathrm{fast}}$ (\cref{sec:eval_protocol}) in points per 0.1\,m/s on all 80 tasks, with 95\% intervals from resampling tasks.}
\label{tab:supp-peak-slopes}
\footnotesize
\setlength{\tabcolsep}{2pt}
\begin{tabular}{llcc}
\toprule
Model & Input & $\beta_{\mathrm{slow}}$ & $\beta_{\mathrm{fast}}$ \\
\midrule
$\pi_{0.5}$ & Current & +23.7\,{\scriptsize[+19.2, +28.3]} & $-$20.8\,{\scriptsize[$-$24.3, $-$17.1]} \\
 & +Past & +22.9\,{\scriptsize[+18.6, +27.2]} & $-$25.4\,{\scriptsize[$-$28.7, $-$21.8]} \\
\addlinespace[1pt]
$\pi_0$-FAST & Current & +15.0\,{\scriptsize[+11.2, +19.1]} & $-$10.8\,{\scriptsize[$-$14.0, $-$7.7]} \\
 & +Past & +18.1\,{\scriptsize[+14.3, +22.2]} & $-$16.1\,{\scriptsize[$-$19.1, $-$13.1]} \\
\addlinespace[1pt]
OpenVLA-OFT & Current & +17.4\,{\scriptsize[+12.1, +22.7]} & $-$11.4\,{\scriptsize[$-$14.5, $-$8.5]} \\
 & +Past & +18.6\,{\scriptsize[+13.4, +23.6]} & $-$10.6\,{\scriptsize[$-$13.4, $-$7.9]} \\
\addlinespace[1pt]
GR00T N1.7 & Current & +11.5\,{\scriptsize[+7.8, +15.6]} & $-$10.6\,{\scriptsize[$-$13.0, $-$8.2]} \\
 & +Past & +14.7\,{\scriptsize[+11.3, +18.4]} & $-$11.4\,{\scriptsize[$-$14.7, $-$8.3]} \\
\addlinespace[1pt]
Cosmos Policy & Current & +22.1\,{\scriptsize[+17.0, +27.4]} & $-$18.6\,{\scriptsize[$-$23.1, $-$14.4]} \\
 & +Past & +12.6\,{\scriptsize[+7.9, +17.5]} & $-$11.8\,{\scriptsize[$-$15.8, $-$7.9]} \\
\addlinespace[1pt]
VLA-JEPA & Current & +3.2\,{\scriptsize[+0.3, +6.1]} & $-$6.1\,{\scriptsize[$-$7.8, $-$4.6]} \\
 & +Past & +6.8\,{\scriptsize[+4.2, +9.4]} & $-$7.8\,{\scriptsize[$-$10.3, $-$5.6]} \\
\addlinespace[1pt]
DynamicVLA & Current & $-$50.4\,{\scriptsize[$-$56.8, $-$44.0]} & $-$0.5\,{\scriptsize[$-$1.4, 0.0]} \\
 & +Past & $-$1.5\,{\scriptsize[$-$4.8, +1.7]} & $-$20.0\,{\scriptsize[$-$24.6, $-$15.5]} \\
\addlinespace[1pt]
Fast-WAM & Current & $-$37.0\,{\scriptsize[$-$43.2, $-$30.8]} & $-$0.9\,{\scriptsize[$-$2.7, 0.0]} \\
 & +Past & $-$12.7\,{\scriptsize[$-$17.3, $-$8.3]} & $-$3.4\,{\scriptsize[$-$5.8, $-$0.8]} \\
\bottomrule
\end{tabular}
\end{table}

\paragraph{Single tasks.}
With the same 12 policies pooled, 68 of the 80 tasks have their highest success rate at 0.15--0.25\,m/s, and none has it at 0.10\,m/s.
Task--speed pairs that no policy solves do not create the peak either: only 5 of the 1,040 pairs have no success from any of the 16 policies in \cref{tab:speed13}.

\paragraph{Weighting every demonstration equally.}
Slow demonstrations contribute more frames (\cref{tab:supp-demos}).
Fine-tuning $\pi_{0.5}$ with past frames again so that every demonstration is sampled equally often leaves its success rate on the 20-task subset unchanged (37.4\%, against 37.6\% for the original policy), and it still peaks at intermediate speeds (peak score $+23.7$ $[+16.2, +31.0]$ at these seven speeds, against $+20.0$ $[+12.7, +27.2]$).

\paragraph{Speed-dependent success tolerances.}
In 53 tasks, the tolerance on how closely the object must follow the belt widens with belt speed, so a slow belt is judged more strictly.
Fixing it at its 0.25\,m/s value in all 53 tasks changes the success rate of $\pi_{0.5}$ with past frames by $-0.1$ $[-1.3, +0.9]$ points, and the policy still peaks at intermediate speeds (peak score $+23.5$ $[+17.7, +28.4]$).

\subsection{Static counterparts}
\label{sec:supp-static-results}
\begin{figure}[t]
  \centering
  \includegraphics[width=\linewidth]{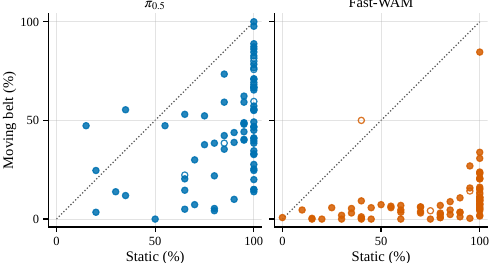}
  \caption{\textbf{Static against moving-belt success per task} for the current-frame policies (one dot per task; static: 20 episodes, moving: 13 speeds $\times$ 20 episodes).
  Hollow: the eight tasks of \cref{tab:supp-static} whose static counterpart changes the task itself (all but Bead pour and Surface wipe). Dotted: equal success.}
  \label{fig:supp-static-moving}
\end{figure}

To check that the tasks can be learned when the object does not move, we fine-tune $\pi_{0.5}$, Fast-WAM, DynamicVLA and VLA-JEPA with the current frame on the static counterparts.
On the 20-task subset they succeed in 87.2\% ($\pi_{0.5}$), 70.8\% (Fast-WAM), 65.5\% (DynamicVLA), and 19.5\% (VLA-JEPA) of episodes.
For $\pi_{0.5}$ and Fast-WAM we also check all 80 tasks: they succeed in 86.9\% and 76.6\% of episodes, against 46.1\% and 9.5\% on the moving belt.
Most moving-belt failures, 71\% for $\pi_{0.5}$ and 61\% for Fast-WAM, occur on tasks with at least 80\% static success (\cref{fig:supp-static-moving}).
Conversely, no task that succeeds in fewer than half of its static episodes exceeds 60\% on the moving belt, for either policy.
The static-counterpart policies train for the same number of epochs on longer demonstrations, so for about 1.8 times as many steps, and the gap also reflects this longer training.

\subsection{Gains from past frames}
\begin{figure*}[t]
  \centering
  \includegraphics[width=\textwidth]{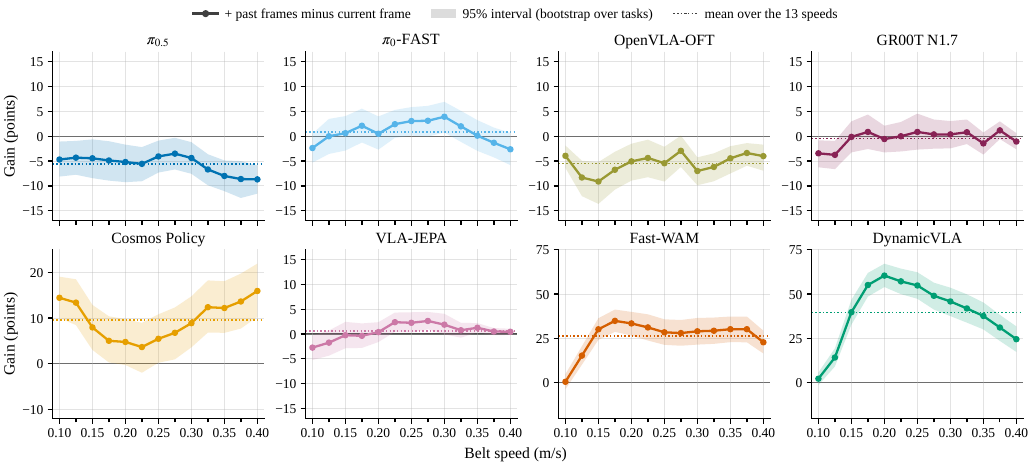}
  \caption{\textbf{Gain in success rate from past frames at each speed} (percentage points; 80 tasks).
  Bands: 95\% intervals from resampling tasks; dotted: mean over the 13 speeds.
  Cosmos Policy, Fast-WAM, and DynamicVLA use larger y-scales.}
  \label{fig:supp-gain-speed}
\end{figure*}

\cref{tab:supp-gains} gives each model's gain with its interval, and \cref{fig:supp-gain-speed} the gain at each speed.
\begin{table}[t]
  \centering
  \scriptsize
  \setlength{\tabcolsep}{2.3pt}
  \caption{\textbf{Gain from past frames} (percentage points; mean over 80 tasks and 13 speeds, with 95\% intervals from resampling tasks).}
  \label{tab:supp-gains}
  \begin{tabular}{@{}lr@{}}
    \toprule
    Model & Gain \\
    \midrule
    $\pi_{0.5}$   & $-5.6$ $[-7.4, -3.9]$ \\
    $\pi_0$-FAST  & $+0.9$ $[-0.5, +2.2]$ \\
    OpenVLA-OFT   & $-5.5$ $[-7.8, -3.5]$ \\
    GR00T N1.7    & $-0.5$ $[-1.9, +1.0]$ \\
    Cosmos Policy & $+9.6$ $[+5.8, +13.2]$ \\
    VLA-JEPA      & $+0.6$ $[-0.1, +1.2]$ \\
    DynamicVLA    & $+39.5$ $[+34.2, +44.8]$ \\
    Fast-WAM      & $+26.4$ $[+21.0, +32.0]$ \\
    \bottomrule
  \end{tabular}
\end{table}

\paragraph{Past frames against the ground-truth speed.}
A large gain from past frames may also only reflect a weak current-frame policy.
We therefore divide the success rate of each past-frame policy by that of the same model given the ground-truth speed (\cref{sec:told}), on the 20-task subset at seven speeds: the past-frame policy reaches 76\% of it for DynamicVLA, 60\% for Fast-WAM, 52\% for $\pi_{0.5}$, and 49\% for VLA-JEPA.

\subsection{End-effector actions for \texorpdfstring{$\pi_{0.5}$}{pi0.5}}
\label{sec:supp-ee}
The two models that gain most from past frames, Fast-WAM and DynamicVLA, are the only ones that act in end-effector space; the others act in joint space.
A $\pi_{0.5}$ fine-tuned with end-effector states and actions succeeds as often as the joint-space one on the 40-task subset, with the current frame (43.4\% vs.\ 43.8\% in joint space) and with past frames (39.8\% vs.\ 38.8\%).
End-effector actions alone therefore do not make $\pi_{0.5}$ gain from past frames.

\subsection{Accelerating belt}
\label{sec:supp-accel-results}
At $a = 0$, the policies fine-tuned on accelerating belts succeed as often as the constant-speed ones or more often ($+3.5$ to $+13.9$ points), so training on acceleration costs no success on a constant 0.25\,m/s belt (\cref{fig:accel} shows only changes relative to $a = 0$).
Their success rates at $a = 0$ are 66.9\% for $\pi_{0.5}$, 38.0\% for Fast-WAM, 21.9\% for VLA-JEPA, and 65.9\% for DynamicVLA, against 53.0\%, 34.5\%, 16.9\%, and 61.4\% for the constant-speed policies (800 episodes each).
At $a = +0.05$\,m/s$^2$, the belt reaches 0.25\,m/s only when the object passes the robot, so the policies meet the object on a slower belt, and part of each change from $a = 0$ would also occur on a constant belt at that speed.
We estimate this part from each policy's success rates at constant speeds, interpolated at the belt speed where the object passes closest to the gripper.
After subtracting this part, the change of the accelerating-belt policy is $+7.9$ $[+3.3, +12.7]$ points instead of $+14.4$ for Fast-WAM and $-13.2$ $[-19.6, -7.0]$ instead of $-30.8$ for $\pi_{0.5}$, and is indistinguishable from zero for DynamicVLA and VLA-JEPA.

\subsection{Failure modes}
\label{sec:supp-qual}
\begin{figure*}[t]
  \centering
  \includegraphics[width=\textwidth]{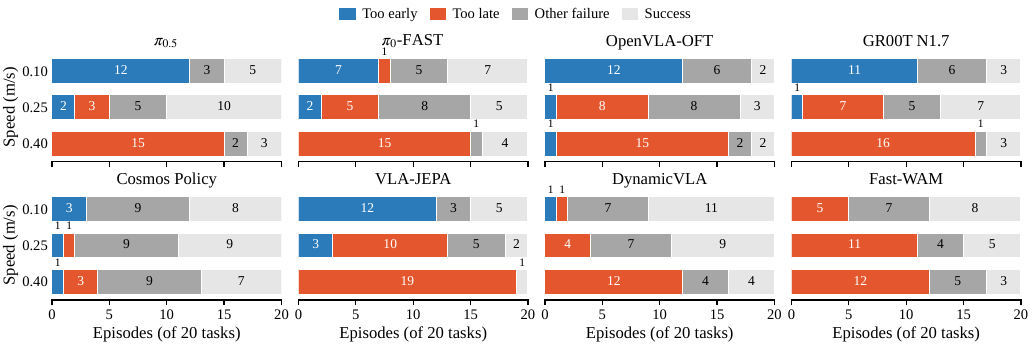}
  \caption{\textbf{Hand-labeled outcomes per policy} (the first evaluation episode of each task in the 20-task subset, past frames; 20 episodes per bar).}
  \label{fig:supp-qual-failures}
\end{figure*}

For the eight past-frame policies, we labeled by hand the 354 failures among 480 episodes: the first evaluation episode of each task in the 20-task subset at 0.10, 0.25, and 0.40\,m/s (\cref{fig:supp-qual-failures}).
We labeled the episodes one at a time in random order, without knowing their policy.
As in \cref{sec:curves}, a failure is too early if the policy acts before the object arrives and too late if it acts after the object has passed; acting before the part the policy must meet arrives also counts as too early, and failing to catch up with the object as too late.
The figures draw the remaining failures as \emph{other failure}.
At 0.10\,m/s, 57 of the 58 too-early failures come from the six models that peak at intermediate speeds, and at 0.40\,m/s too-late failures are the most common class for every model except Cosmos Policy.

\section{Probes}
\label{sec:supp-probe}
\begin{figure*}[t]
  \centering
  \includegraphics[width=\textwidth]{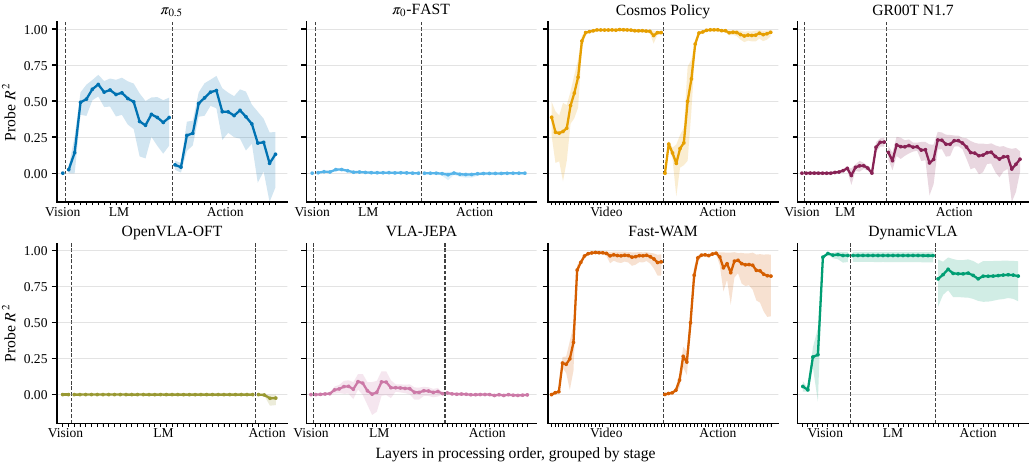}
  \caption{\textbf{Held-out probe $R^2$ of the past-frame policies at every layer}, in processing order; dashed lines separate the vision or video model, the language model, and the action module; bands: 95\% intervals from resampling tasks.}
  \label{fig:supp-probe-layers}
\end{figure*}

This section gives the protocol of the probes in \cref{sec:probe}; \cref{fig:supp-probe-layers} shows the probe $R^2$ at every layer of the past-frame policies.

\paragraph{Data.}
We record 5 episodes per task and speed at all 13 speeds on the 12 tasks of the 20-task subset that do not start with an object in the hand.
The arm is held at its start pose while the object passes, which keeps the policy's own speed-dependent motion out of the features.
From each episode we take the time steps at which the object reaches 8 positions along the belt, the same at every speed, so that the object's position does not reveal the speed (6,240 samples).

\paragraph{Probes.}
At every layer of each model, we average the frozen features over the tokens of the current exterior image and fit a ridge regression to the belt speed.
For GR00T N1.7's causal language model we average over the tokens after the last image instead, and for action modules over their action tokens at the first denoising step.
GR00T N1.7's language-model stage also includes the four self-attention blocks at the start of its action head, and VLA-JEPA's action module is read at the action head's input, over its 32 action-query tokens.
We report held-out $R^2$ from 5-fold cross-validation split by task; the ridge penalty and the layer of each stage are chosen by inner cross-validation on the training tasks of each fold.
Choosing instead the layer with the highest held-out $R^2$ would raise each past-frame policy's highest value in \cref{fig:probe} by at most 0.07.

\paragraph{Baselines.}
Current-frame policies give $R^2 \le 0$ at every layer, and probes of the past-frame policies refit on speeds shuffled within each task stay below 0.02.
As references that use the object's true position, projected from the simulator into the cameras, the same ridge regression reaches $R^2 = 0.54$ on the optical flow at that position and 0.72 on the object's pixel displacement across the past frames.
\paragraph{Pretrained weights.}
For the pretrained weights in \cref{sec:probe}, each value is the highest held-out $R^2$ over all layers, so high values are slightly optimistic and low ones are upper bounds.
GR00T N1.7's pretrained action module is not probed: its state and action encoders are specific to each robot, and the pretrained weights have none for our arm.

\begin{figure}[t]
  \centering
  \includegraphics[width=\linewidth]{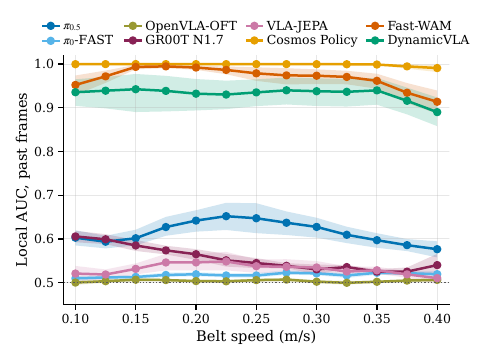}
  \caption{\textbf{Probe quality at each belt speed.} Local AUC of each past-frame policy: how well the probe orders samples at neighboring speeds (dotted: 0.5, the value of random guessing, where every current-frame policy lies); bands: 95\% intervals from resampling tasks.}
  \label{fig:supp-probe-per-speed}
\end{figure}

\paragraph{Probe quality at each speed.}
$R^2$ is undefined within one speed, so at each speed we use a local AUC (\cref{fig:supp-probe-per-speed}): for this speed and a neighboring one, the probability that the probe predicts a higher speed for a sample at the faster of the two than for a sample of the same task at the slower one, averaged over the 12 tasks and the speed's one or two neighbors.
Here the inner cross-validation chooses among all of a model's read points, from any stage, and its predicted actions; it never chooses the predicted actions for a past-frame policy.
At the high speeds where their success rate falls, the probes of Cosmos Policy, Fast-WAM, and DynamicVLA still separate neighboring speeds, so these failures are unlikely to come from the speed no longer being decodable.

\section{Adapter}
\label{sec:supp-adapter}
\cref{tab:adapter} gives the full comparison behind \cref{sec:adapter} at the four speeds where the adapter is evaluated, and \cref{sec:supp-real} describes the adapter refit for the real robot.
\begin{table}[t]
\centering
\caption{\textbf{Feeding an estimated speed to a speed-conditioned policy.}
Success rate (\%) on the 20-task subset at 0.10, 0.20, 0.30, and 0.40\,m/s (20 episodes per task and speed), all with past frames.
\emph{None}: the past-frame policy without speed input; \emph{Adapter} and \emph{GT}: the speed-conditioned policy given the adapter's estimate or the ground-truth speed.
\emph{Recovered} (\%): (Adapter $-$ None) / (GT $-$ None); \emph{Diff.}: Adapter minus GT, in points; brackets: 95\% intervals from resampling tasks.}
\label{tab:adapter}
\footnotesize
\setlength{\tabcolsep}{1.6pt}
\begin{tabular}{lccccc}
\toprule
Model & None & Adapter & GT & Recovered & Diff. \\
\midrule
$\pi_{0.5}$  & 32.9 & 69.8 & 70.6 & 98 {\scriptsize[91, 105]} & $-0.9$ {\scriptsize[$-3.3$, $+1.6$]} \\
Fast-WAM     & 30.1 & 48.3 & 48.6 & 98 {\scriptsize[87, 111]} & $-0.3$ {\scriptsize[$-2.2$, $+1.8$]} \\
VLA-JEPA     & 11.3 & 23.7 & 23.1 & 105 {\scriptsize[88, 128]} & $+0.6$ {\scriptsize[$-1.4$, $+2.7$]} \\
DynamicVLA   & 46.6 & 59.4 & 61.9 & 83 {\scriptsize[61, 100]} & $-2.6$ {\scriptsize[$-5.3$, $-0.1$]} \\
\bottomrule
\end{tabular}
\end{table}

\paragraph{Estimator.}
A frozen RAFT-small model computes the optical flow between two exterior images 0.1\,s apart.
From the flow and the pixel coordinates, a convolutional head with 2,946 parameters predicts a speed and a score for each of $14 \times 14$ image cells ($16 \times 16$ pixels each); the estimate is the mean of the cell speeds, weighted by the softmax of the scores over the cells.

\paragraph{Training.}
We train the head on ten frame pairs from each of 640 training demonstrations (8 per task) and test it on 160 held-out demonstrations (2 per task).
On these, the estimate from a single frame pair has a mean absolute error of 0.010\,m/s.
All demonstrations share the benchmark's single belt texture, so the adapter is not tested on unseen textures.

\paragraph{In closed loop.}
Before control step 2, the policy is given a prior of 0.25\,m/s.
From step 2 on, the adapter makes one estimate per step from the frames at $t-2$ and $t$, and the policy is given the mean of the latest five (fewer before step 6); this mean has a mean absolute error of 0.007\,m/s.

\paragraph{Re-planning at the first estimate.}
The first action chunk is planned at step 0 with the prior.
In all simulated adapter runs, the policy drops this chunk and plans a new one as soon as the first estimate exists (step 2); the policies given the ground-truth speed or no speed keep the schedule of \cref{alg:supp-exec}.
Without this re-planning, $\pi_{0.5}$ with the adapter succeeds in 62.1\% instead of 69.8\% of episodes over the four speeds and recovers about 77\% instead of 98\% of its gain from the ground-truth speed.
Re-planning may also favor the adapter: frames before the episode start repeat the first frame (Fast-WAM masks them; \cref{alg:supp-exec}), so the chunk planned at step 2 is the first one whose past frames show motion.

\FloatBarrier
\section{Real Robot}
\label{sec:supp-real}
\begin{figure}[t]
  \centering
  \includegraphics[width=\columnwidth]{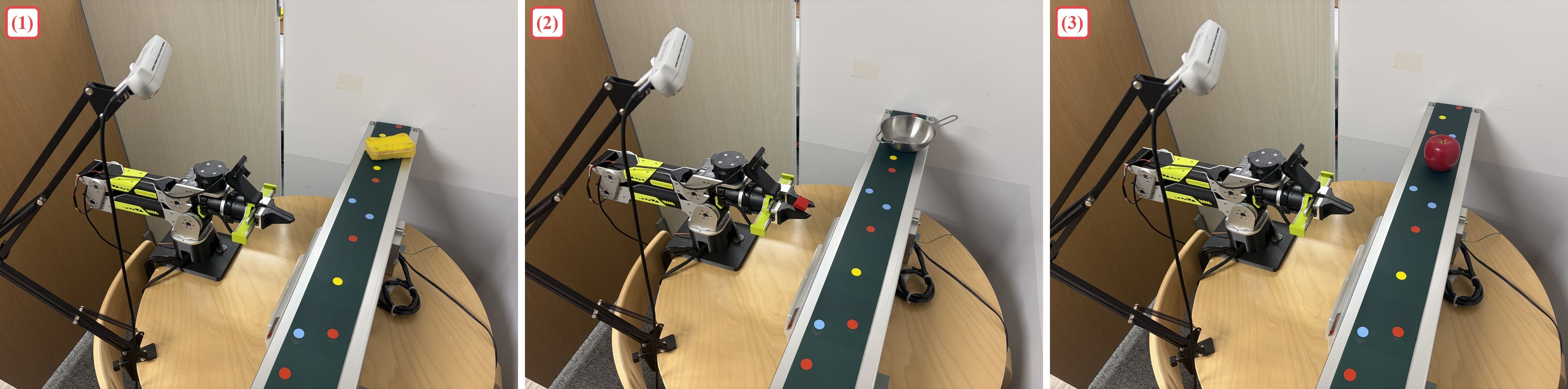}
  \caption{\textbf{The three real-robot tasks} at the start of a trial: (1) ``Pick the yellow sponge'', with the sponge on the belt; (2) ``Place the red cube in the plate'', with the cube in the gripper and the plate on the belt; (3) ``Push the apple off the belt''.
  A trial is completed if the sponge is grasped, the cube rests on the plate, or the apple falls off the belt.}
  \label{fig:supp-real-tasks}
\end{figure}

\begin{figure}[t]
  \centering
  \includegraphics[width=\columnwidth]{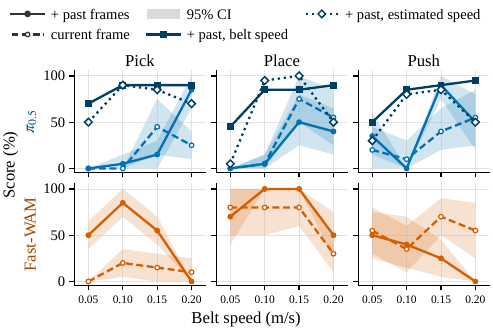}
  \caption{\textbf{Real-robot score vs.\ belt speed per task.}
  Each point is the mean score of 10 trials; bands: 95\% bootstrap intervals over trials, for the policies without speed input.
  \cref{fig:real-results} shows the mean over the three tasks.}
  \label{fig:supp-real-per-task}
\end{figure}

\cref{fig:supp-real-tasks} shows the three tasks, \cref{tab:real} gives the scores behind \cref{sec:real}, and \cref{fig:supp-real-per-task} gives the score of each task at each speed.
\begin{table}[t]
\centering
\caption{\textbf{Real-robot score (\%)} of $\pi_{0.5}$ and Fast-WAM with the current frame (Cur.) or two past frames (+Past), and of $\pi_{0.5}$ with past frames given the belt speed set for the trial (+Past, GT) or the adapter's estimate (+Past, est.); 10 trials per task and speed. \emph{All}: the three tasks at 0.05--0.20\,m/s. \emph{0.10--0.20}: the mean over 0.10, 0.15, and 0.20\,m/s, the speeds that both cover, on the robot (Real) and in simulation (Sim.: success rate of the policies of \cref{tab:speed13} on all 80 tasks; unlike these, the real-robot $\pi_{0.5}$ uses openpi's default configuration).}
\label{tab:real}
\small
\setlength{\tabcolsep}{2pt}
\begin{tabular}{llrrrrrr}
\toprule
& & \multicolumn{4}{c}{0.05--0.20\,m/s} & \multicolumn{2}{c}{0.10--0.20\,m/s} \\
\cmidrule(lr){3-6}\cmidrule(lr){7-8}
Model & Input & Pick & Place & Push & All & Real & Sim. \\
\midrule
$\pi_{0.5}$ & Cur. & 17.5 & 33.8 & 31.2 & 27.5 & 34.4 & 43.2 \\
 & +Past & 26.2 & 23.8 & 43.8 & 31.2 & 37.8 & 38.5 \\
 & +Past, GT & 85.0 & 76.2 & 80.0 & 80.4 & 88.9 & -- \\
 & +Past, est. & 73.8 & 62.5 & 61.2 & 65.8 & 78.3 & -- \\
Fast-WAM & Cur. & 11.2 & 67.5 & 53.8 & 44.2 & 43.9 & 22.9 \\
 & +Past & 47.5 & 80.0 & 28.8 & 52.1 & 50.6 & 44.2 \\
\bottomrule
\end{tabular}
\end{table}

\paragraph{Data and fine-tuning.}
The belt speeds (0.05--0.20\,m/s) are lower than in simulation because, unlike the simulator, the robot keeps moving while a remote GPU computes the next actions (Execution below).
Each policy uses its settings in \cref{tab:supp-recipes} ($\pi_{0.5}$: the real-robot column), with the number of steps set to give each task as many training samples as in simulation ($\pi_{0.5}$: as with openpi's default configuration, 30,000 steps at batch size 32): 1,125 steps for $\pi_{0.5}$ and 369 for Fast-WAM.
The past frames are the head-camera images from 0.1 and 0.2\,s earlier, the same offsets as in simulation.

\paragraph{Speed input.}
The adapter of \cref{sec:supp-adapter} is retrained on the 108 training demonstrations and gives the policy the mean of its latest five estimates.
In all 120 trials with the estimate, its trial mean is within 0.025\,m/s of the set speed.

\paragraph{Execution.}
With the latency-aware action streaming of \cref{sec:real}, each new action chunk drops the actions for the control steps that have passed since its observation.

\paragraph{Past frames.}
The real-robot $p$-values, here and in \cref{sec:real}, come from permutation tests that shuffle the input labels among the 20 trials (10 per policy) of each task and speed.
For Fast-WAM the effect of past frames depends on the task: its score rises from 11 to 48 on picking ($p<0.001$) and falls from 54 to 29 on pushing ($p=0.01$).
The difference between Fast-WAM's and $\pi_{0.5}$'s gains from past frames is $+4.2$ $[-7.1, +15.0]$ points over the three tasks (95\% interval from resampling the 10 trials of each policy, input, task, and speed), $+27.5$ on picking, $+22.5$ on placing, and $-37.5$ on pushing.

\begin{table*}[p]
\centering
\caption{\textbf{The 80 tasks, grouped by skill category.} Italics mark the words of the instruction that change between episodes; the number in parentheses after an instruction counts the distinct instructions in the task's demonstrations (after a category, its number of tasks). Objects: number of object instances in the task's object set. Expert: the scripted expert's success rate (\%) at 0.10, 0.20, 0.30, and 0.40\,m/s, over 384 episodes per task and speed. Diagnostic tasks follow the object without touching it. $^\dagger$: 20-task subset; $^\dagger$ or $^\ddagger$: 40-task subset; $^\S$: accelerating-belt tasks.}
\label{tab:supp-tasks}
\scriptsize
\setlength{\tabcolsep}{3pt}
\begin{tabular}{@{}p{3.0cm}p{10.0cm}rrrrr@{}}
\toprule
 & & & \multicolumn{4}{c@{}}{Expert (\%) at speed (m/s)} \tabularnewline
\cmidrule(l){4-7}
Task & Instruction & Objects & 0.1 & 0.2 & 0.3 & 0.4 \tabularnewline
\midrule
\multicolumn{7}{@{}l}{\textbf{Place} (17)} \tabularnewline
\raggedright Arch-clearance load$^\S$ & \raggedright put the \emph{red box} in the pallet bin and pull your hand clear of the arch (6) & 10 & 100 & 100 & 99 & 98 \tabularnewline
\raggedright Belt pick-and-place$^\dagger$ & \raggedright pick up the \emph{blue box} from the conveyor belt and place it in the bin (83) & 602 & 92 & 92 & 97 & 96 \tabularnewline
\raggedright Corner belt transfer$^\S$ & \raggedright move the \emph{green can} from the main conveyor onto the branch conveyor (8) & 16 & 92 & 93 & 93 & 92 \tabularnewline
\raggedright Counter-flow belt swap & \raggedright move the \emph{red box} from the far conveyor belt onto the near conveyor belt (6) & 8 & 100 & 100 & 100 & 100 \tabularnewline
\raggedright Even two-pad load & \raggedright lay the bar evenly across both pads on the conveyor belt & 1 & 100 & 100 & 100 & 97 \tabularnewline
\raggedright Flawed-part cull$^\dagger$$^\S$ & \raggedright inspect the \emph{green can} on the conveyor belt: if it carries a defect band, put it in the reject bin; otherwise leave it alone and let it ride past (4) & 5 & 100 & 99 & 100 & 100 \tabularnewline
\raggedright Fruit drop$^\dagger$ & \raggedright put the \emph{apple} into the red bowl on the conveyor belt (6) & 6 & 100 & 100 & 100 & 100 \tabularnewline
\raggedright Gentle set-down$^\S$ & \raggedright set the \emph{green box} down gently on the pad on the conveyor belt (4) & 4 & 100 & 98 & 99 & 100 \tabularnewline
\raggedright Kit-cell fill$^\dagger$ & \raggedright put the \emph{yellow} box into the empty cell of the kit tray on the conveyor belt (4) & 4 & 100 & 98 & 99 & 100 \tabularnewline
\raggedright Named-patch cover$^\ddagger$$^\S$ & \raggedright \emph{cover the blue patch on the board with the gray tile, leave the red one showing} (4) & 6 & 100 & 100 & 100 & 100 \tabularnewline
\raggedright Overhang recenter$^\ddagger$$^\S$ & \raggedright move the overhanging \emph{blue box} back onto the middle of the conveyor belt (4) & 5 & 92 & 92 & 92 & 91 \tabularnewline
\raggedright Rail balance$^\S$ & \raggedright balance the \emph{red box} on the rail on the conveyor belt (3) & 4 & 100 & 100 & 99 & 100 \tabularnewline
\raggedright Ramp roll & \raggedright release the \emph{teal} ball at the top of the ramp so it rolls into the catch pocket (2) & 2 & 100 & 100 & 100 & 100 \tabularnewline
\raggedright Reader presentation$^\ddagger$ & \raggedright hold the \emph{yellow-marked} face of the part up to the reader window, then set it back on the belt (4) & 4 & 100 & 100 & 100 & 100 \tabularnewline
\raggedright Relative place$^\ddagger$$^\S$ & \raggedright place the \emph{blue box in front of} the marker block on the tray on the conveyor belt (16) & 12 & 100 & 100 & 100 & 100 \tabularnewline
\raggedright Topping onto cake$^\ddagger$$^\S$ & \raggedright put the \emph{red} box inside the printed target zone on the cake on the conveyor belt (4) & 4 & 96 & 96 & 98 & 96 \tabularnewline
\raggedright Tray-to-bin transfer$^\S$ & \raggedright move the \emph{green box} from the tray into the bin on the conveyor belt (8) & 12 & 100 & 100 & 100 & 100 \tabularnewline
\midrule
\multicolumn{7}{@{}l}{\textbf{Pick} (14)} \tabularnewline
\raggedright Belt grasp & \raggedright pick up the \emph{blue box} from the conveyor belt (24) & 63 & 85 & 87 & 89 & 88 \tabularnewline
\raggedright Crushable-shell pick & \raggedright pick up the crushable shell from the conveyor belt without squashing it & 1 & 97 & 100 & 100 & 99 \tabularnewline
\raggedright Handle grasp$^\dagger$$^\S$ & \raggedright pick up the \emph{basket} by its handle (4) & 4 & 100 & 100 & 100 & 96 \tabularnewline
\raggedright Hanging-load settle & \raggedright lift the carry frame off the conveyor belt and steady its hanging weight & 1 & 100 & 99 & 90 & 93 \tabularnewline
\raggedright Keyed-part lift$^\S$ & \raggedright slide the \emph{orange} part along its slot and lift it off the carrier (4) & 4 & 100 & 99 & 98 & 98 \tabularnewline
\raggedright Loaded-tray lift$^\S$ & \raggedright lift the \emph{green} tray off the conveyor belt without spilling the ball (4) & 12 & 100 & 100 & 100 & 100 \tabularnewline
\raggedright Named-object pick$^\dagger$ & \raggedright pick up the \emph{blue box} from the conveyor belt (9) & 25 & 100 & 100 & 100 & 100 \tabularnewline
\raggedright Ordinal pick & \raggedright \emph{grab the first block to arrive on the conveyor belt} (6) & 1 & 100 & 100 & 100 & 100 \tabularnewline
\raggedright Packed-cluster pick$^\ddagger$ & \raggedright lift a part out of the packed tray on the conveyor belt & 1 & 95 & 97 & 95 & 86 \tabularnewline
\raggedright Plowed-lane pick$^\dagger$$^\S$ & \raggedright pick up the block as the plough pushes it across the conveyor belt & 1 & 100 & 100 & 96 & 92 \tabularnewline
\raggedright Plug-lid lift$^\ddagger$ & \raggedright take the lid off the bin on the conveyor belt & 1 & 100 & 98 & 90 & 96 \tabularnewline
\raggedright Rotating-table pick$^\ddagger$$^\S$ & \raggedright pick the block off the turning table on the conveyor belt & 1 & 99 & 100 & 100 & 98 \tabularnewline
\raggedright Sheet-edge pick & \raggedright lift the framed sheet off its rack by the frame edge without touching the sheet face & 1 & 92 & 92 & 98 & 99 \tabularnewline
\raggedright Tote unload$^\ddagger$ & \raggedright take the \emph{blue box} out of the tote on the conveyor belt (8) & 12 & 100 & 100 & 100 & 100 \tabularnewline
\midrule
\multicolumn{7}{@{}l}{\textbf{Insert} (10)} \tabularnewline
\raggedright Axle seat$^\ddagger$$^\S$ & \raggedright seat the rod in the notched carrier on the conveyor belt & 1 & 100 & 100 & 100 & 99 \tabularnewline
\raggedright Depth-metered insert$^\dagger$$^\S$ & \raggedright insert the rod until the painted band meets the bore mouth & 1 & 100 & 96 & 95 & 98 \tabularnewline
\raggedright Divider-gap lower$^\dagger$$^\S$ & \raggedright lower the \emph{yellow} box into the slot of the crate without touching the dividers (4) & 4 & 100 & 100 & 100 & 99 \tabularnewline
\raggedright Dual-pin register$^\ddagger$ & \raggedright fit the bracket onto both pins on the conveyor belt & 1 & 100 & 100 & 100 & 100 \tabularnewline
\raggedright Flap-bin post$^\S$ & \raggedright post the \emph{blue box} through the flap of the bin on the conveyor belt (8) & 8 & 100 & 100 & 99 & 95 \tabularnewline
\raggedright Flush-depth fit & \raggedright seat the \emph{blue box} in the well on the conveyor carrier where its top sits flush (4) & 6 & 100 & 100 & 100 & 100 \tabularnewline
\raggedright Hex-bit alignment$^\S$ & \raggedright seat the hex bit into the hex socket on the conveyor belt & 1 & 98 & 91 & 98 & 97 \tabularnewline
\raggedright Shim under clip$^\dagger$$^\S$ & \raggedright slide the shim under the clip on the conveyor belt & 1 & 100 & 100 & 100 & 99 \tabularnewline
\raggedright Shrouded oblique bore & \raggedright insert the rod into the angled port under the shroud on the conveyor belt & 1 & 100 & 100 & 91 & 96 \tabularnewline
\raggedright Sideways keyway insert & \raggedright insert the rod into the side keyway of the block on the conveyor belt & 1 & 100 & 100 & 99 & 100 \tabularnewline
\midrule
\multicolumn{7}{@{}l}{\textbf{Press} (8)} \tabularnewline
\raggedright Button press$^\ddagger$$^\S$ & \raggedright press the button on the conveyor belt & 1 & 98 & 99 & 100 & 100 \tabularnewline
\raggedright Cap press-fit$^\ddagger$ & \raggedright press the cap onto the jar on the conveyor belt & 1 & 100 & 97 & 97 & 92 \tabularnewline
\raggedright Color-named press$^\S$ & \raggedright press the \emph{red} button on the conveyor belt (4) & 1 & 98 & 97 & 100 & 100 \tabularnewline
\raggedright Label press & \raggedright press the label pad onto the printed patch on the side of the box on the conveyor belt & 1 & 98 & 96 & 100 & 98 \tabularnewline
\raggedright Rocker toggle & \raggedright flip the raised end of the rocker switch on the conveyor belt & 1 & 97 & 97 & 99 & 99 \tabularnewline
\raggedright Sequenced press$^\dagger$$^\S$ & \raggedright \emph{press the red button, then the blue button} (2) & 1 & 89 & 97 & 99 & 96 \tabularnewline
\raggedright Snap-dome stamp$^\S$ & \raggedright stamp the seal pad on the conveyor belt & 1 & 100 & 100 & 100 & 100 \tabularnewline
\raggedright Tare then load$^\dagger$ & \raggedright press the tare button, then put the \emph{yellow} box on the scale pan (3) & 3 & 100 & 99 & 100 & 100 \tabularnewline
\midrule
\multicolumn{7}{@{}l}{\textbf{Rotate} (6)} \tabularnewline
\raggedright Angle match$^\ddagger$ & \raggedright put the puck in the round cradle with its arrow pointing the same way as the display arrow & 1 & 100 & 100 & 100 & 100 \tabularnewline
\raggedright Bottle upright$^\dagger$ & \raggedright stand the fallen \emph{yellow jar} upright on the conveyor belt (4) & 4 & 100 & 100 & 100 & 100 \tabularnewline
\raggedright Commanded face up$^\S$ & \raggedright set the die down on the pad with its \emph{blue} face up (6) & 1 & 100 & 97 & 99 & 95 \tabularnewline
\raggedright Needle set & \raggedright turn the gauge needle \emph{30 degrees clockwise} on the conveyor belt (77) & 1 & 84 & 88 & 88 & 89 \tabularnewline
\bottomrule
\end{tabular}
\end{table*}
\begin{table*}[p]
\centering
{\small\textbf{Table~\ref{tab:supp-tasks}} (continued).}\par\vspace{4pt}
\scriptsize
\setlength{\tabcolsep}{3pt}
\begin{tabular}{@{}p{3.0cm}p{10.0cm}rrrrr@{}}
\toprule
 & & & \multicolumn{4}{c@{}}{Expert (\%) at speed (m/s)} \tabularnewline
\cmidrule(l){4-7}
Task & Instruction & Objects & 0.1 & 0.2 & 0.3 & 0.4 \tabularnewline
\midrule
\multicolumn{7}{@{}l}{\textbf{Rotate} (continued)} \tabularnewline
\raggedright Sole-down place$^\S$ & \raggedright set the \emph{rust} shoe down flat on its sole on the mat on the conveyor belt (3) & 3 & 96 & 96 & 99 & 100 \tabularnewline
\raggedright Yaw-keyed place$^\dagger$$^\S$ & \raggedright place the block into the matching slot on the conveyor belt & 1 & 100 & 95 & 93 & 97 \tabularnewline
\midrule
\multicolumn{7}{@{}l}{\textbf{Push} (5)} \tabularnewline
\raggedright Block topple$^\ddagger$$^\S$ & \raggedright knock over the standing block on the conveyor belt & 1 & 100 & 100 & 100 & 100 \tabularnewline
\raggedright Cover swipe-off$^\S$ & \raggedright swipe the cap off the block on the conveyor belt & 1 & 100 & 100 & 100 & 100 \tabularnewline
\raggedright Drawer close$^\dagger$ & \raggedright push the drawer shut on the conveyor belt & 1 & 100 & 99 & 100 & 84 \tabularnewline
\raggedright Selector set & \raggedright set the selector to the \emph{blue} position on the conveyor belt (3) & 1 & 100 & 94 & 98 & 97 \tabularnewline
\raggedright Slab squaring$^\ddagger$ & \raggedright push the \emph{red box} flush against the tray's rail on the conveyor belt (4) & 7 & 100 & 100 & 98 & 98 \tabularnewline
\midrule
\multicolumn{7}{@{}l}{\textbf{Pull} (4)} \tabularnewline
\raggedright Drawer open$^\S$ & \raggedright slide the drawer tray open on the conveyor belt and let go & 1 & 96 & 100 & 100 & 99 \tabularnewline
\raggedright Peg pull$^\ddagger$ & \raggedright pull the peg out of the block on the conveyor belt & 1 & 92 & 91 & 92 & 95 \tabularnewline
\raggedright Retaining-pin pull & \raggedright pull the retaining pin out of the carrier on the conveyor belt & 1 & 90 & 89 & 90 & 83 \tabularnewline
\raggedright Starter-cord pull & \raggedright pull the starter handle on the conveyor and let it snap back & 1 & 100 & 100 & 100 & 100 \tabularnewline
\midrule
\multicolumn{7}{@{}l}{\textbf{Hang} (3)} \tabularnewline
\raggedright Saddle on rail$^\dagger$ & \raggedright hang the saddle on the rack's rail on the conveyor belt & 1 & 100 & 99 & 100 & 98 \tabularnewline
\raggedright Washer on leaning post$^\ddagger$$^\S$ & \raggedright hang the ring on the tilted post on the conveyor belt & 1 & 98 & 98 & 100 & 100 \tabularnewline
\raggedright Washer on post$^\S$ & \raggedright hang the ring on the post on the conveyor belt & 1 & 99 & 99 & 98 & 98 \tabularnewline
\midrule
\multicolumn{7}{@{}l}{\textbf{Hold} (3)} \tabularnewline
\raggedright Cart coupling hold$^\dagger$$^\S$ & \raggedright hold the front cart until the following cart couples to it, then let the pair ride on & 1 & 100 & 100 & 100 & 100 \tabularnewline
\raggedright Depth-band hold & \raggedright press the plunger down to the painted band and hold it there & 1 & 100 & 100 & 99 & 98 \tabularnewline
\raggedright Spring-tab hold$^\S$ & \raggedright pull the tab out to the marked band and hold it there & 1 & 100 & 100 & 100 & 98 \tabularnewline
\midrule
\multicolumn{7}{@{}l}{\textbf{Pour} (2)} \tabularnewline
\raggedright Bead pour$^\ddagger$$^\S$ & \raggedright pour the beads from the jug into the moving hopper & 1 & 98 & 100 & 99 & 99 \tabularnewline
\raggedright Cup decant & \raggedright tip the ball out of the \emph{stone} cup into the bin without touching the ball (3) & 6 & 95 & 89 & 95 & 100 \tabularnewline
\midrule
\multicolumn{7}{@{}l}{\textbf{Reach} (2)} \tabularnewline
\raggedright Moving-target reach (diagnostic) & \raggedright reach and hold the gripper above the marker on the conveyor belt & 1 & 100 & 100 & 100 & 100 \tabularnewline
\raggedright Scanner aim (diagnostic)$^\dagger$$^\S$ & \raggedright hold the scanner aimed at the code patch on the conveyor belt without touching it & 1 & 100 & 100 & 100 & 100 \tabularnewline
\midrule
\multicolumn{7}{@{}l}{\textbf{Tool use} (2)} \tabularnewline
\raggedright Ball scoop$^\dagger$ & \raggedright scoop the rolling \emph{red} ball off the conveyor belt (3) & 3 & 98 & 99 & 99 & 99 \tabularnewline
\raggedright Ring pass-through$^\S$ & \raggedright pass the rod down through the ring and back out without touching it & 1 & 100 & 100 & 95 & 94 \tabularnewline
\midrule
\multicolumn{7}{@{}l}{\textbf{Open} (1)} \tabularnewline
\raggedright Lid flip & \raggedright flip open the lid of the box on the conveyor belt & 1 & 82 & 85 & 91 & 95 \tabularnewline
\midrule
\multicolumn{7}{@{}l}{\textbf{Close} (1)} \tabularnewline
\raggedright Cabinet-door close$^\dagger$ & \raggedright swing the cabinet door shut on the conveyor belt & 1 & 100 & 99 & 86 & 92 \tabularnewline
\midrule
\multicolumn{7}{@{}l}{\textbf{Screw} (1)} \tabularnewline
\raggedright Wing-nut run-down$^\S$ & \raggedright screw the wing nut down the post until it seats on the conveyor belt & 1 & 100 & 100 & 99 & 92 \tabularnewline
\midrule
\multicolumn{7}{@{}l}{\textbf{Wipe} (1)} \tabularnewline
\raggedright Surface wipe$^\ddagger$$^\S$ & \raggedright wipe the marks off the passing board & 1 & 100 & 100 & 96 & 98 \tabularnewline
\bottomrule
\end{tabular}
\end{table*}


\FloatBarrier
\section{Licenses and Release}
\label{sec:supp-repro}
We will release the benchmark code and task definitions under the MIT license, and the demonstrations under CC BY-NC 4.0 because their images include ReplicaCAD backgrounds.
The backgrounds come from ReplicaCAD~\cite{szot2021habitat2} (CC BY-NC 4.0), RoboCasa~\cite{nasiriany2024robocasa} (CC BY 4.0), and ArchitecTHOR~\cite{deitke2022procthor} (Apache 2.0); the object meshes from Google Scanned Objects~\cite{downs2022gso}, YCB~\cite{calli2015ycb}, and RoboCasa (CC BY 4.0) and from Poly Haven (CC0), or are generated by our code; and the platform materials from Poly Haven and ambientCG (both CC0).

\end{document}